\documentclass[a4paper, journal]{IEEEtran}
\pdfoutput=1 
\usepackage{siunitx}                
\usepackage{threeparttable}         
\usepackage{booktabs}               
\usepackage{adjustbox}              
\usepackage{breqn}                  
\usepackage{tabularx}
\usepackage[acronym]{glossaries-extra}  
\usepackage{glossary-mcols}             
\usepackage[en-GB]{datetime2}           
\usepackage{mathtools}                  
\usepackage{xfrac}                      
\usepackage{accents}                    
\usepackage[inline]{enumitem}           
\usepackage[
    bottom,
    hang,
    flushmargin]{footmisc}              

\usepackage[dvipsnames]{xcolor}     
\usepackage{tikz}                   
\usepackage{graphicx}
\usepackage{microtype}
\usepackage{subcaption}
\usepackage{transparent,import}     
\usepackage{placeins}               
\usepackage{epstopdf}
\usepackage[colorlinks=true]{hyperref}
\usepackage[
    backend=biber,
    style=ieee,
    citestyle=numeric-comp,
    mincitenames=1,
    maxcitenames=1,
    maxbibnames=5,
    url=false,
    doi=true]
           {biblatex}
\usepackage[
 textwidth=2.75cm]{todonotes}    
\DeclareNameAlias{author}{family-given}
\DeclareNameAlias{editor}{family-given}
\DeclareNameAlias{translator}{family-given}
\newcommand*{\citet}{\textcite}
\newcommand*{\citep}{\parencite}

\newcolumntype{C}{>{\centering\arraybackslash}X} 
\AtBeginDocument{\sisetup{math-rm=\mathrm, detect-weight=true, binary-units = true, mode=text,range-phrase = --, range-units=single}}
\DeclareSIUnit\degree{deg}
\DeclareSIUnit\usd{USD}
\DeclareSIUnit\rpm{rpm}
\DeclareSIUnit\gauss{G}
\DeclareSIUnit\pixel{px}
\DeclareSIUnit\revolution{rev}
\DeclareSIUnit\pp{p.p.}
\newsavebox{\largestimage}
\DTMusemodule{british}{en-GB}

\usepackage{lscape,pdflscape}
\usepackage{amsmath,amssymb,bm,amsfonts,amsthm,eucal}
\usepackage{multirow}
\usepackage{graphicx}

\usepackage{url}
\usepackage{makecell}

\usepackage{epsfig}

\def\Figref#1{Figure~\ref{#1}}
\def\figref#1{Fig.~\ref{#1}}
\def\Secref#1{Section~\ref{#1}}
\def\secref#1{Sec.~\ref{#1}}
\def\Subsecref#1{Subsection~\ref{#1}}

\def\acrpos#1{\glsxtrlong*{#1}'s (\glsxtrshort{#1})\glsunset{#1}}
\def\acrconnect#1#2{%
\ifglsused{#1}
    {\glsxtrshort{#1}{#2}%
    }
    {\glsxtrshort{#1}{#2} (\glsxtrlong*{#1})\glsunset{#1}%
    }%
}
\def\acrcite#1#2{%
\ifglsused{#1}
    {\glsxtrshort{#1} \parencite{#2}%
    }
    {\glsxtrlong{#1} (\glsxtrshort*{#1} \cite{#2})\glsunset{#1}%
    }%
}

\def\acrsquare#1{%
\ifglsused{#1}
    {\glsxtrshort{#1}%
    }
    {\glsxtrlong{#1} [\glsxtrshort*{#1}]\glsunset{#1}%
    }%
}

\def\figref#1{Fig.~\ref{#1}}
\def\Figref#1{Figure \ref{#1}}

\def\eqref#1{Eq.~(\ref{#1})}

\def\Tabref#1{Table \ref{#1}}

\def\vq{{\bm{q}}}
\def\vr{{\bm{r}}}

\def\vt{{\bm{t}}}

\def\mR{{\bm{R}}}

\def\mT{{\bm{T}}}

\DeclareMathAlphabet{\mathsfit}{\encodingdefault}{\sfdefault}{m}{sl}
\SetMathAlphabet{\mathsfit}{bold}{\encodingdefault}{\sfdefault}{bx}{n}

\def\urr{{\textnormal{r}}}

\def\urt{{\textnormal{t}}}

\def\gL{{\mathcal{L}}}

\DeclareRobustCommand\undervec[1]{\underaccent{\vec}{#1}}

\def\Fb{\undervec{\bm{\mathcal{F}}}_{b}}
\def\Fc{\undervec{\bm{\mathcal{F}}}_{c}}

\def\Fi{\undervec{\bm{\mathcal{F}}}_{i}}
\def\Fo{\undervec{\bm{\mathcal{F}}}_{o}}
\def\Fs{\undervec{\bm{\mathcal{F}}}_{s}}
\def\Ft{\undervec{\bm{\mathcal{F}}}_{t}}

\def\Fpi{\undervec{\bm{\mathcal{F}}}_{\Pi}}

\newcommand{\normltwo}{L^2}

\ExplSyntaxOn

\NewDocumentCommand \irow { s o m }
 {
  \IfBooleanTF {#1}
   { \vectaux*{#3} }
   { \IfValueTF {#2} { \vectaux[#2]{#3} } { \vectaux{#3} } }
 }

\DeclarePairedDelimiterX \vectaux [1] {\lbrack} {\rbrack}
 { \, \dbacc_vect:n { #1 } \, }

\cs_new_protected:Npn \dbacc_vect:n #1
 {
  \seq_set_split:Nnn \l_tmpa_seq { , } { #1 }
  \seq_use:Nn \l_tmpa_seq { \enspace }
 }
\ExplSyntaxOff

\makeglossaries
\setabbreviationstyle[acronym]{long-short}
\glssetwidest{ARCADEEE}
\newacronym{ai}{AI}{artificial intelligence}
\newacronym{amds}{AMDS}{Autonomous Micro-satellite Docking System}
\newacronym{atv}{ATV}{Automated Transfer Vehicle}
\newacronym{arcade}{ARCADE}{Autonomous Rendezvous Control and Docking Experiment}
\newacronym{ard}{ARD}{Autonomous Rendezvous and Docking}
\newacronym{asds}{ASDS}{Autonomous Satellite Docking System}
\newacronym{assist}{ASSIST}{Harmonized System Study on Interfaces and Standardization of Fuel Transfer}
\newacronym{asmil}{ASMIL}{Autonomous Systems and Machine Intelligence Laboratory}

\newacronym{bilstm}{BiLSTM}{bi-directional long-short-term memory}

\newacronym{cad}{CAD}{computer aided design}
\newacronym{cnc}{CNC}{computer numerical control}
\newacronym{cnn}{CNN}{convolutional neural network}

\newacronym[shortplural={DOF},longplural=degrees-of-freedom]{dof}{DOF}{degree-of-freedom}
\newacronym{dr}{DR}{design requirement}
\newacronym{dl}{DL}{deep learning}
\newacronym{dnn}{DNN}{deep neural network}
\newacronym{drcnn}{DRCNN}{deep recurrent convolutional neural network}

\newacronym{fr}{FR}{functional requirement}
\newacronym{fov}{FOV}{field of view}
\newacronym{fc}{FC}{fully connected}
\newacronym{gap}{GAP}{global average pooling}
\newacronym{gpu}{GPU}{graphics processing unit}

\newacronym{ip}{IP}{image processing}
\newacronym{iss}{ISS}{International Space Station}

\newacronym[longplural=long-short-term memories]{lstm}{LSTM}{long-short-term memory}

\newacronym{mev}{MEV}{Mission Extension Vehicle}
\newacronym{ml}{ML}{machine learning}

\newacronym{oibar}{OIBAR}{Orbital AI-based Autonomous Refuelling}
\newacronym{oos}{OOS}{on-orbit servicing}
\newacronym{or}{OR}{operational requirement}

\newacronym{pr}{PR}{performance Requirement}
\newacronym{pnp}{P$n$P}{perspective-$n$-point}
\newacronym{pi}{PI}{proportional integral}

\newacronym{rnn}{RNN}{recurrent neural network}
\newacronym{rgb}{RGB}{red-green-blue}
\newacronym{rv}{RV}{rendezvous}
\newacronym{rvdb}{RVD/B}{rendezvous and docking or berthing}

\newacronym{sv}{SV}{service vehicle}
\newacronym{sva}{SVA}{service vehicle arm}
\newacronym{spec}{SPEC}{Spacecraft Pose Estimation Challenge}

\newacronym{tle}{TLE}{two-line element}
\newacronym{trl}{TRL}{technology readiness level}
\newacronym{tv}{TV}{target vehicle}

\newacronym{udp}{UDP}{Universal Docking Port}

\newacronym{vbs}{VBS}{vision-based sensor}

\begin{document}

\title{Orbital AI-based Autonomous Refuelling Solution}

\author{Duarte~Rondao,
        Lei~He,
        and~Nabil~Aouf
\thanks{D. Rondao is a Postdoctoral Research Fellow with the Department of Electrical and Electronic Engineering at City, University of London, EC1V 0HB, UK (e-mail: \texttt{duarte.rondao@city.ac.uk}).}%
\thanks{L. He was a Postdoctoral Research Fellow with the Department of Electrical and Electronic Engineering at City, University of London, EC1V 0HB, UK (currently a PhD candidate at Northwestern Polytechnical University, Xi’an, China).}%
\thanks{N. Aouf is a Professor of Robotics and Autonomous Systems with the Department of Electrical and Electronic Engineering at City, University of London, EC1V 0HB, UK.}%
}

\markboth{DRAFT MANUSCRIPT}%
{Rondao \MakeLowercase{\textit{et al.}}: OIBAR: Orbital AI-based Autonomous Refuelling}

\maketitle

\begin{abstract}

Cameras are rapidly becoming the choice for on-board sensors towards space rendezvous due to their small form factor and inexpensive power, mass, and volume costs. When it comes to docking, however, they typically serve a secondary role, whereas the main work is done by active sensors such as lidar. This paper documents the development of a proposed \acrconnect{ai}{-based} navigation algorithm intending to mature the use of on-board visible wavelength cameras as a main sensor for docking and \gls{oos}, reducing the dependency on lidar and greatly reducing costs. Specifically, the use of \gls{ai} enables the expansion of the relative navigation solution towards multiple classes of scenarios, e.g., in terms of targets or illumination conditions, which would otherwise have to be crafted on a case-by-case manner using classical image processing methods. Multiple \gls{cnn} backbone architectures are benchmarked on synthetically generated data of docking manoeuvres with the \gls{iss}, achieving position and attitude estimates close to \SI{1}{\percent} range-normalised and \SI{1}{\degree}, respectively. The integration of the solution with a physical prototype of the refuelling mechanism is validated in laboratory using a robotic arm to simulate a berthing procedure.

\end{abstract}

\printglossary[type=\acronymtype,title={List of Acronyms},style=alttree,nonumberlist]
\glsresetall

\IEEEpeerreviewmaketitle

\section{Introduction}

\noindent For the majority of the \num{64}-year history of space launches, satellites have been seen as an expendable medium: once the propellant is depleted, the mission is ended. Northrop Grumman’s \gls{mev} programme has recently challenged this paradigm by achieving the first teleoperated \gls{oos} to reposition existing spacecraft. This has opened up a new market segment where the spacecraft’s life cycle can be extended beyond its original planning, avoiding the costs of launching, manufacturing, and keeping a new one. The success of the mission has attracted the attention of the United States Department of Defence, which have awarded the company a contract to study the possibility of servicing commercial and government satellites using robotics technology \cite{northrop2020}.

Despite its breakthrough, it is incontrovertible that the \gls{mev} was built on the shoulders of previous demonstrators: for example, the Kepler \acrpos{atv} first refuelling operation of the \gls{iss} to supply the station’s thrusters in \num{2011} \cite{benarroche2014atv}, or NASA's Robotic Refuelling Mission which demonstrated the technology to refuel satellites in orbit by robotic means in \num{2014} \citep{metcalfe2014robotic}. More recently, ESA have also recognised the potential for this new market segment by opening up calls for ideas related to \gls{oos}, having previously invested \SI{50}[M€]{} in support for research and development of relevant technologies \cite{esa2021}. Overall, \gls{oos} and manufacturing alone is projected to have a cumulative global market size of over \SI{4.4}[B\$]{} by \num{2030}, of which it is predicted that the UK could capture \SI{1}[B\$]{} \citep{uksa2021}. Still, existing satellites were, and still are, built without thinking of their serviceability and, more specifically, refuelling, which is a fulcral part of the servicing operations of these assets and represents a significant cost saving measure.

Paramount to the safe accomplishment of refuelling, and \gls{oos} in general, is the estimation of the relative states between the \gls{sv} and the client or \gls{tv} during docking or berthing. At such small distances, this entails the estimation of the six \gls{dof} relative pose, which is typically achieved with two types of optical sensors: lidar and camera sensors  \citep{wie2014attitude}. However, current flight-proven solutions using either sensor require optical corner-cube reflectors to be mounted on the \gls{tv} \citep{fehse2003sensors}. The viability of large-scale \gls{oos} involves \gls{rvdb} via autonomous navigation with minimal or no human input, and the cost associated with active sensors can hinder the massification of orbital servicers. Indeed, cameras are already rapidly becoming the choice for on-board sensors towards spacecraft \gls{rv} due to their small form factor and inexpensive power, mass, and volume costs. The past three years have witnessed a consolidation of \acrconnect{ai}{-based} techniques for \gls{rv}, particularly through \gls{dl}, using monocular cameras which do not make assumptions about the level of cooperation of the target \cite{song2022deep}. However, this has not yet been established as a navigation approach for docking.

\section{Related Work}
\label{sec:relwork}

\begin{figure}[t]
	\centering
	\includegraphics[width=\columnwidth]{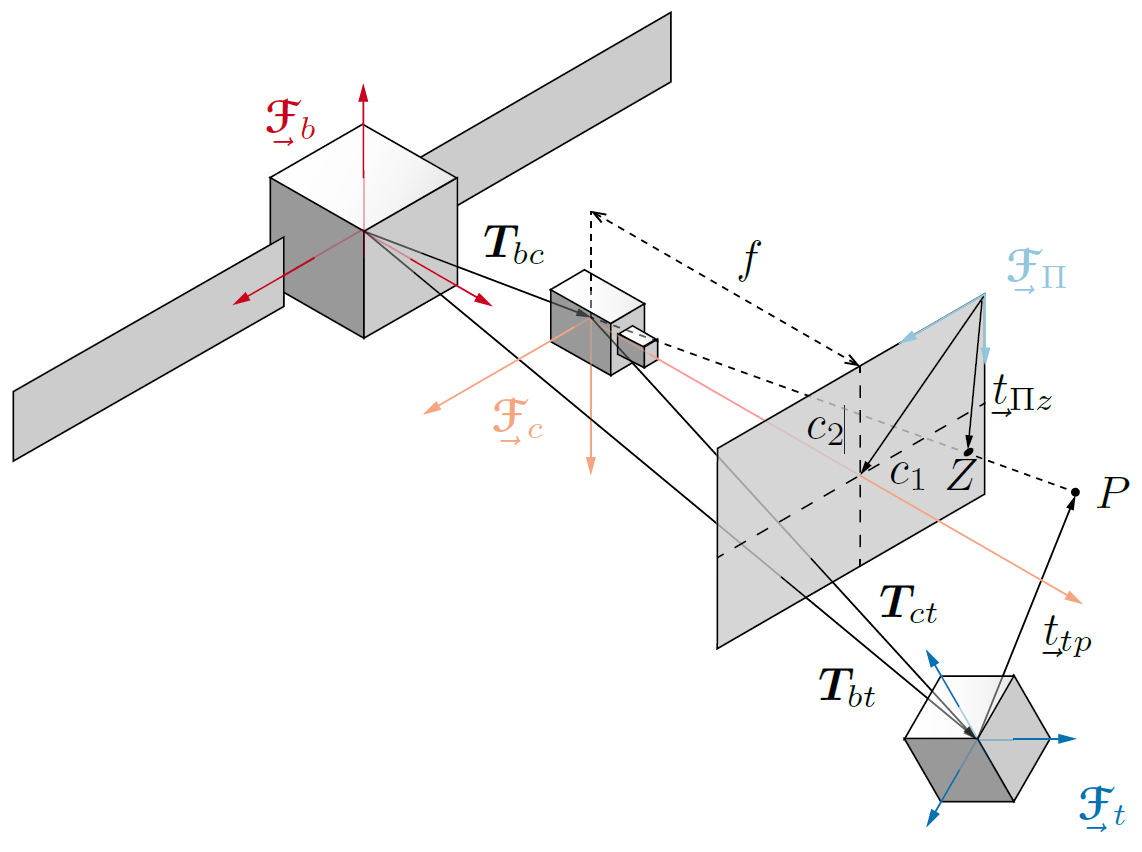}
 \caption{Frames of reference involved in the relative pose estimation problem for \gls{rvdb} \citep{rondao2021multimodal}.}
\label{fig:relwork-bodyframes}
\end{figure}

\noindent The use of \glspl{vbs} for \gls{rvdb} has traditionally consisted of establishing geometric relationships derived from the laws of imaging on the focal plane of a lens \cite{fehse2003sensors}: the \gls{sv} illuminates retro-reflectors on the \acrconnect{tv}{-side}, which are configured according to a known pattern and are then imaged by the on-board camera. By knowing this configuration (i.e., the relative distances between the pattern markers), and the intrinsic parameters of a calibrated \gls{vbs} (i.e., the \acrsquare{fov} and focal length), information on the \gls{sv}-\gls{tv} range, line of sight direction, and relative attitude can be computed by detecting said markers on each image. In the context of \gls{rvdb}, navigation requires the estimation of said quantities, which make up the 6-\gls{dof} relative pose $\mT_{bt}$ mapping the target vehicle frame of reference $\Ft$ to the service vehicle frame $\Fb$ (\figref{fig:relwork-bodyframes}). This entails the need for relative navigation sensors which, in the case of a \gls{vbs}, define two extra frames: the physical camera frame $\Fc$ (which can often be assumed coincident with $\Fb$ without loss of generality) and the image plane frame $\Fpi$ containing the image of the \gls{tv} and where the \gls{ip} tasks occur.

In the computer vision literature, this problem is called the \acrcite{pnp}{szeliski2022computer}. \gls{pnp} can be used with only four markers to retrieve the full relative pose, although additional markers can be used to robustify the solution. \acrconnect{pnp}{-based} pose estimation, while accurate, can be defeated somewhat easily if the marker detection pipeline is not reliable enough. In a space \gls{rvdb} scenario, for example, using traditional \gls{ip} techniques for detection can falter under the rapidly changing lighting conditions a \gls{vbs} is subjected to. Furthermore, reliance on retro-reflective markers restricts the navigation algorithm to cooperative targets, limiting its range of applications.

In contrast, modern \gls{ai} techniques, namely through the advent of \gls{dl} and \glspl{dnn}, have gained resurgence in the field of computer vision from the beginning of the previous decade onward, due to advances in commercial-off-the-shelf \glspl{gpu} and accessibility to large-scale datasets such as ImageNet \cite{deng2009imagenet}. The eruption in popularity of \gls{dl} arguably occurred in \num{2012} with AlexNet \cite{krizhevsky2017imagenet}, a \gls{cnn} which won the ImageNet Large Scale Visual Recognition Challenge with a top-5 classification error more than almost \num{11} percent points lower than the runner-up; the novel use of \acrconnect{gpu}{-based} training massively accelerated the process, enabling deep learning to be competitive. \glspl{cnn}, i.e., \glspl{dnn} tailored to process image inputs through efficient convolution kernels, later became the norm, as new designs have competed in the challenge each year, resulting in exponential classification score improvements. Two notable examples are GoogLeNet \cite{szegedy2015going}, marked not only by a very deep architecture, but also by the implementation of parallel layers to extract multi-scale features; and ResNet \cite{he2015deep}, which introduced residual connections allowing the breakthrough to even deeper architectures. Most of these state-of-the-art \glspl{cnn} have been open-sourced and made available with pre-trained ImageNet weights, which has since contributed to the swift evolution of \gls{dl} in general and in the adoption of such models as \gls{cnn} front-ends.

It took more than five years for the popularity of \glspl{cnn} to migrate onto the domain of spacecraft relative pose estimation for \gls{rv}. In 2019, ESA Kelvins’ \gls{spec} benchmarked estimation errors obtained on image inputs taken with on-board \gls{vbs} from a simulated \gls{rv} with the Tango spacecraft \cite{kisantal2020satellite}. Although it did not tackle docking or berthing, it did demonstrate the good performance of \acrconnect{ai}{-based} approaches in vision-based navigation for space.

\begin{figure}[t]
    \centering
	\includegraphics[width=\columnwidth]{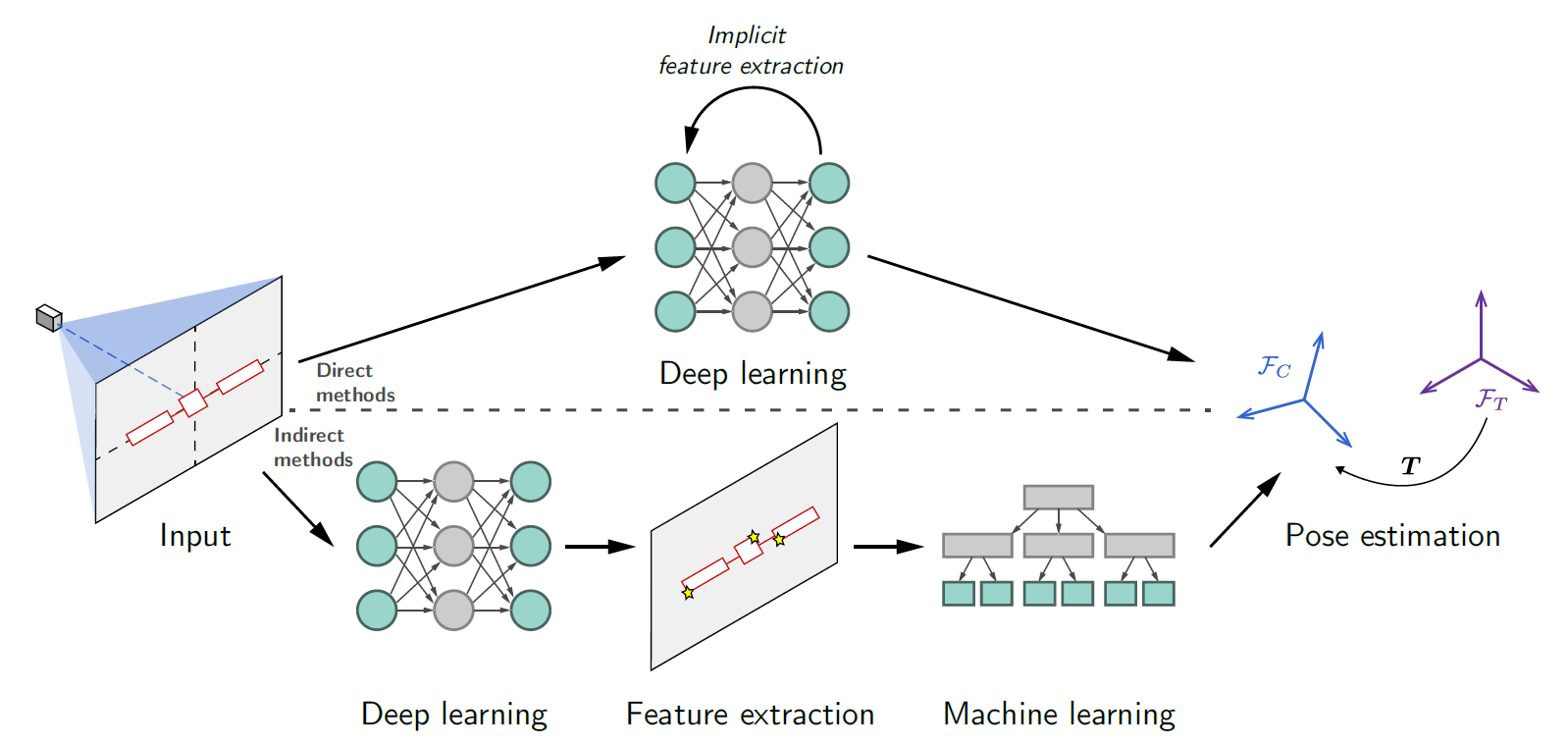}
	\caption{Direct versus indirect methods for DL-based pose estimation \cite{song2022deep}.}
	\label{fig:relwork-rv}
\end{figure}

Interestingly, most of the highest scoring \gls{spec} competitors \cite{song2022deep, chen2019satellite} followed a so-called indirect approach towards \acrconnect{dl}{-based} pose estimation (\figref{fig:relwork-rv}, bottom): this entails relaying the use of a \gls{cnn} entirely to the feature extraction task, and then use a \gls{ml} method, such as a \gls{pnp} solver, to retrieve the solution based on the features. These are pre-defined by the user at the pre-training stage and can be ``natural'' landmarks on the surface of the target, i.e., they are not limited to physical fiducial or retro-reflective markers, making it appropriate for uncooperative scenarios. This type of model-based but uncooperative approach had previously been demonstrated using traditional \gls{ip} techniques \cite{rondao2021robust}, but \acrconnect{dl}{-based} approaches eliminate the need of crafting feature detection and matching techniques that are potentially specific to each different scenario, which is desirable.

The abovementioned techniques, however, have not been evaluated on docking scenarios. Landmark detection traditionally degrades at very close relative proximity, and \acrconnect{dl}{-based} indirect methods are not expected to behave differently. Furthermore, at such distances some of the landmarks may be outside the \gls{fov}, degrading the solution. For the scope of the \gls{oibar} project, a \acrconnect{dl}{-based} direct \gls{vbs} navigation method is instead investigated. This approach (\figref{fig:relwork-rv}, top) involves designing a \gls{dnn} which generates a pose directly from image inputs, i.e., it is end-to-end, and has been tested for close-range rendezvous both in terms of assorted imagery \cite{proencca2020deep} and continuous trajectories \cite{rondao2021chinet}. Such end-to-end methods are desirable as they do not rely on additional \gls{ml} pipelines to generate the pose, and open the door to \acrconnect{dl}{-based} temporal modelling, which has been shown to improve the solution for time-series where there is a correlation between successive relative poses \cite{rondao2021chinet}.
\section{Methodology and Design}

\noindent This section details the approach followed for the execution of the \gls{oibar} project. The design of the docking mechanism has originally been detailed by \citet{lei2023novel}; a rundown of the refuelling operations is reiterated below for context. Then, the \acrconnect{ai}{-based} navigator introduced in this paper is described.

\subsection{Refuelling Operations}

\begin{figure}[t]
  	\centering
  	\begin{subfigure}[t]{0.49\columnwidth}\centering
  		\includegraphics[width=\textwidth]{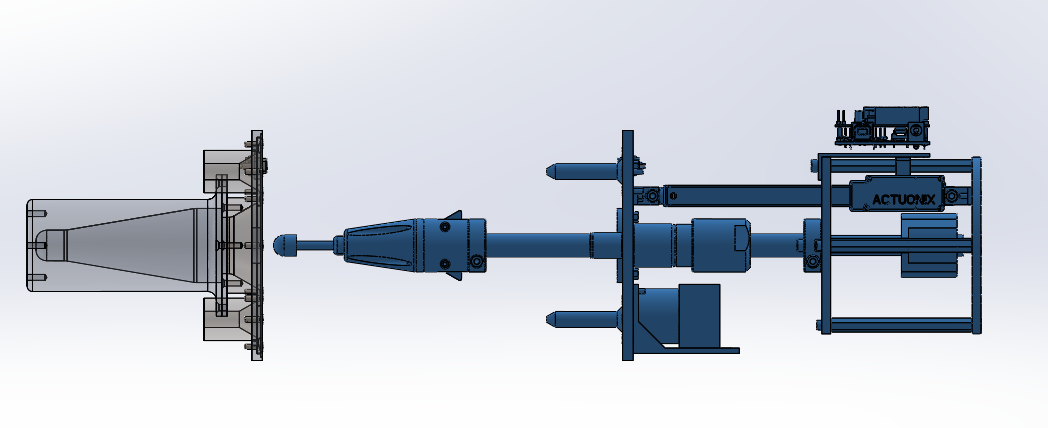}
      		\caption{Approach}
      		\label{fig:method-refuelling1}
  	\end{subfigure}
  	\begin{subfigure}[t]{0.49\columnwidth}\centering
  		\includegraphics[width=\textwidth]{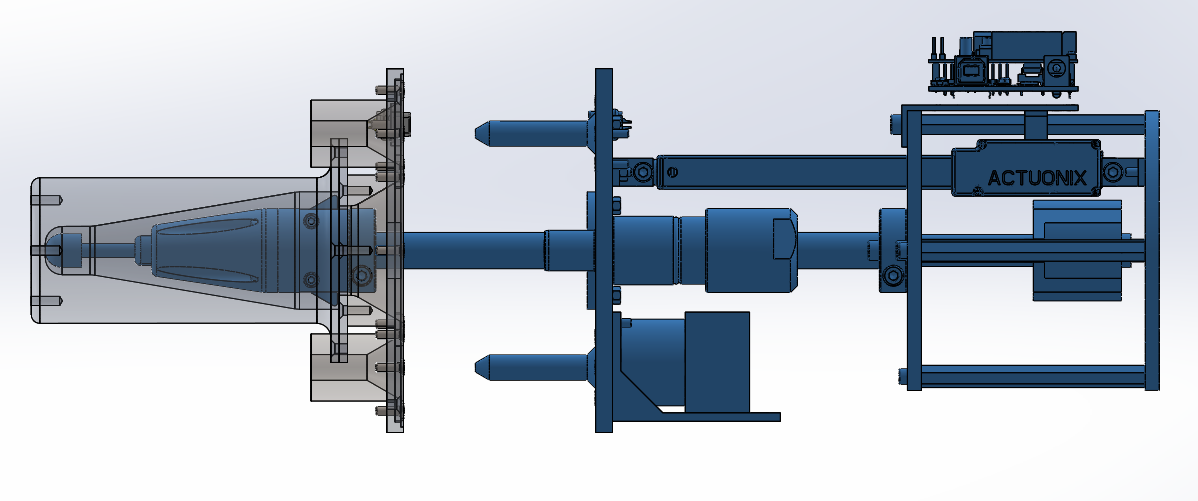}
      		\caption{Reception and soft-docking}
      		\label{fig:method-refuelling2}
  	\end{subfigure}\\
  	\begin{subfigure}[t]{0.49\columnwidth}\centering
  		\includegraphics[width=\textwidth]{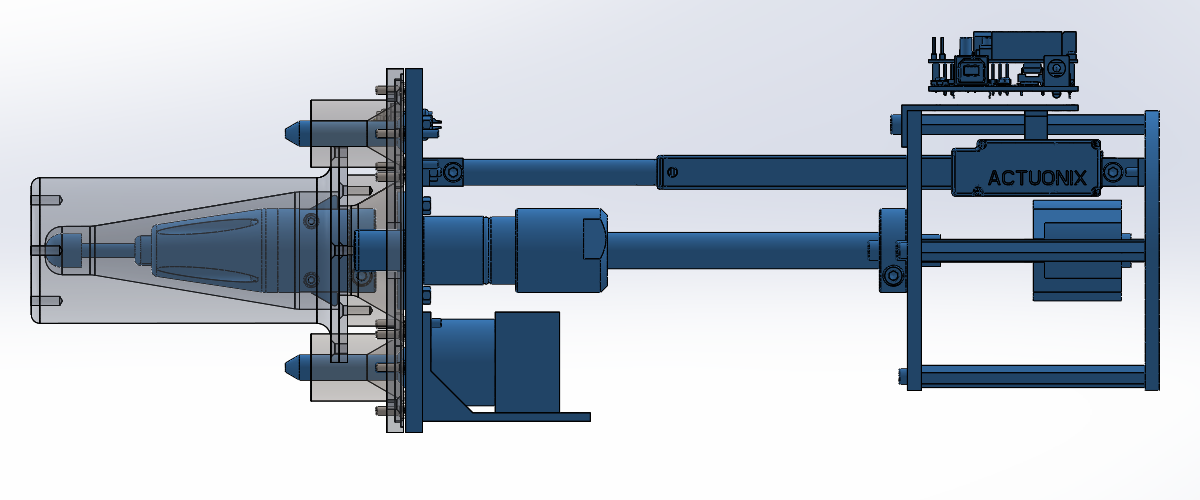}
      		\caption{Hard-docking}
      		\label{fig:method-refuelling3}
  	\end{subfigure}
  	\begin{subfigure}[t]{0.49\columnwidth}\centering
  		\includegraphics[width=\textwidth]{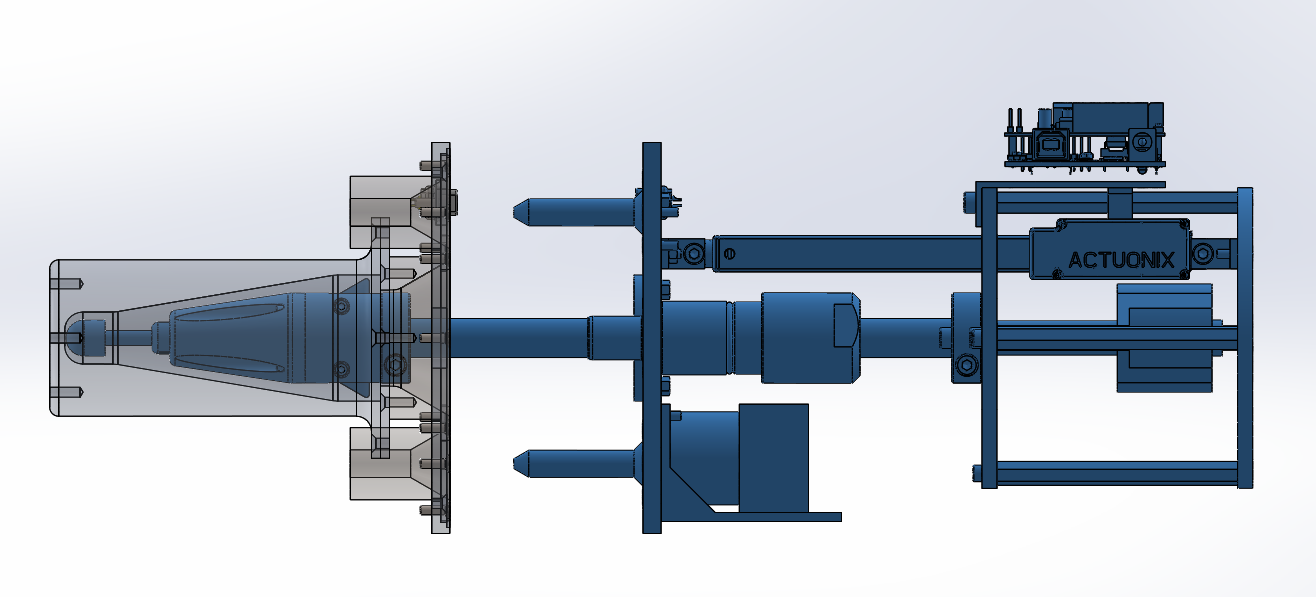}
      		\caption{Fluidic plane disengagement}
      		\label{fig:method-refuelling4}
  	\end{subfigure}\\
  	  	\begin{subfigure}[t]{0.49\columnwidth}\centering
  		\includegraphics[width=\textwidth]{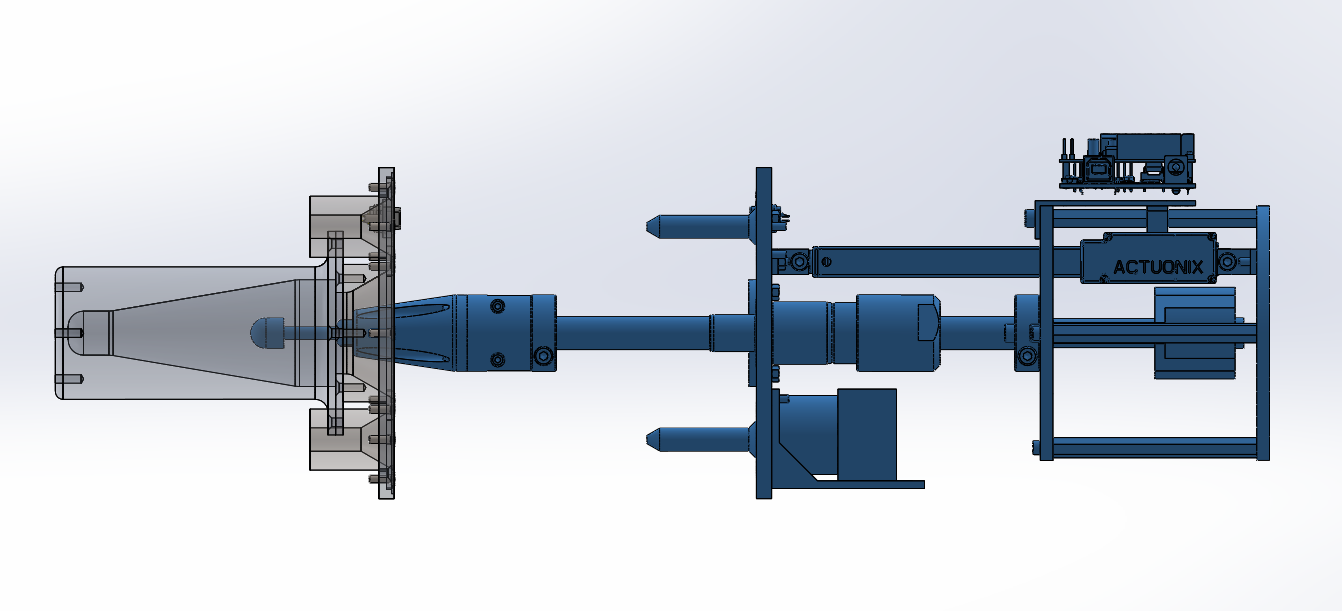}
      		\caption{Separation}
      		\label{fig:method-refuelling5}
  	\end{subfigure}
  \caption{Refuelling operations between \gls{tv} and \gls{sv} using the designed docking mechanism for \gls{oibar} \citep{lei2023novel}.}
  \label{fig:method-refuelling}
\end{figure}

\noindent The refuelling procedure in \gls{oibar} can be broken down into the following sequence of operations, which are also illustrated in \Figref{fig:method-refuelling}.

\subsubsection{Approach}

\noindent In this phase, the \gls{sv} will approach the \gls{tv} guided by the \acrconnect{ai}{-based} navigation algorithm. The approach velocity will be reduced to the final value. The service satellite will achieve the lateral and angular alignment to place each docking interface into the other’s reception range, as shown in \Figref{fig:method-refuelling1}.

\subsubsection{Reception}

\noindent In this phase, the probe of the end-effector enters the drogue of the berthing fixture. The funnel-like design of the drogue cavity will guide the end-effector probe, limiting the misalignment. Within the probe, a spring-damper is designed to absorb the contact shock until it reaches the end of the drogue. This also provide the retraction force for the separation phase after completing the refuelling process.

\subsubsection{Soft-docking}

\noindent The probe of the end-effector is equipped with a soft-docking latch. This includes two spring-loaded latches to achieve the execution of the soft-docking procedure. When the probe moves into the drogue, these latches are pulled back by the wall contact forces, retracting into the probe to allow it to enter the drogue cavity. The drogue is designed to have a varying diameter cross-section; as such, once the latch enters the drogue cavity past a certain point, contact with the wall ceases and the spring-loaded latches are naturally pushed back out. This prevents the end-effector from accidentally exiting the drogue (e.g., due to reaction forces), thus completing the soft-docking phase, as shown in \Figref{fig:method-refuelling2}.

\subsubsection{Hard-docking}

\noindent After soft-docking, the end-effector is restricted by the berthing fixture to ensure sufficient tolerance for the hard-docking. Then, the fluidic plane is pushed towards the berthing fixture until connection is achieved, as illustrated in in \Figref{fig:method-refuelling3}. The alignment pins will engage in the guide cavities on the berthing fixture side. These guides are tapered to eliminate any minor misalignment. Once the fluidic plane is connected to the other half on the berthing fixture, all fluid couplings and electrical connector become, by extension, also connected correctly.

\subsubsection{Fuel Transfer}

\noindent This stage initiates the transfer of fuel through by opening the valves of both \gls{sv} and \gls{tv}. Once the tank pressure reaches the target pressure, the values are closed back. Control commands and the pressure data are communicated through the electrical connector.

\subsubsection{Separation}

\noindent After refuelling, the end-effector is released from the berthing fixture. During this phase, the fluidic plane will be retracted by the actuator firstly, as shown in \Figref{fig:method-refuelling4}. Then, the soft-docking latch will be released by the latch stepper motor. The end-effector will be extracted automatically by relying on the pre-loaded force from the spring-damper. Finally, the latch stepper motor will control the latch to the default position to get ready for the next refuelling process, as shown in \Figref{fig:method-refuelling5}.

\subsection{Software Development}

\subsubsection{Architecture}

\begin{figure*}[t]
    \centering
	\includegraphics[width=\textwidth]{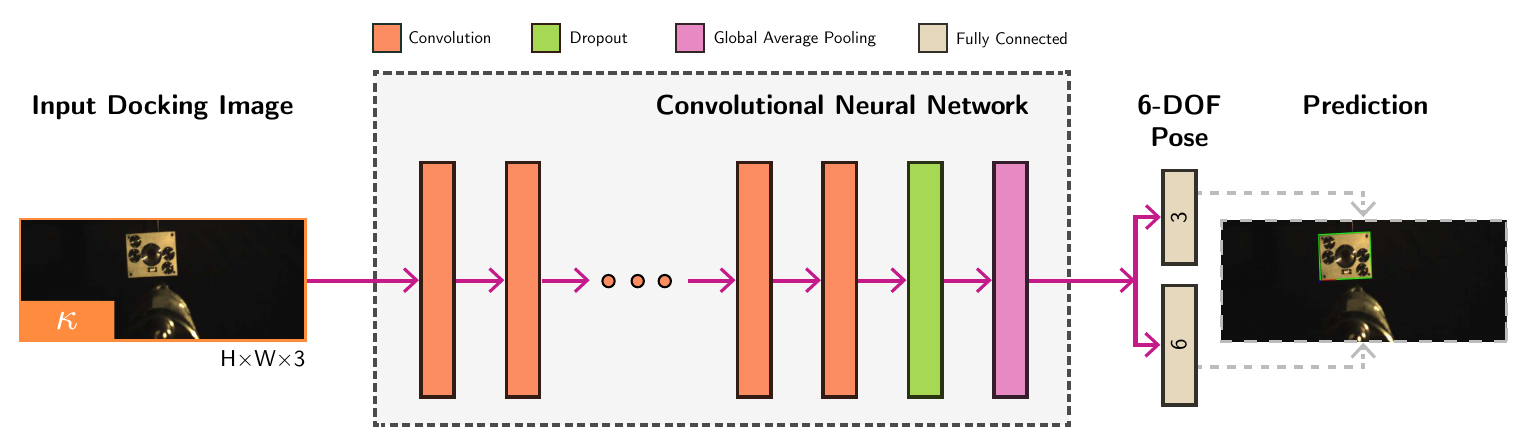}
	\caption{OibarNet base \gls{dnn} architecture for \acrconnect{ai}{-based} docking navigation.}
	\label{fig:method-architecture}
\end{figure*}

\noindent \Figref{fig:method-architecture} illustrates the base architecture of the \acrconnect{ai}{-based} navigation algorithm for \gls{oibar}, termed OibarNet.

The proposed network is a direct (i.e., end-to-end) \gls{dnn} taking \gls{rgb} images of the \gls{tv}'s berthing fixture at time-step $\tau=k$ and outputting the corresponding 6-\gls{dof} pose relative to the \gls{sv}. The front-end and backbone of OibarNet is the \gls{cnn} that processes the image inputs (\figref{fig:method-architecture}, in orange). Multiple \gls{cnn} model candidates are considered and evaluated within the project (see Section 4); however, two architectural aspects are kept constant. The first is a dropout layer (\figref{fig:method-architecture}, in green) added after the last dropout layer to prevent overfitting \cite{hinton2012improving}. The second is a \gls{gap} layer (\figref{fig:method-architecture}, in pink). \gls{gap} converts the \gls{cnn} output to a fixed-dimension vector dependant only on the number of output channels, regardless of the image inputs spatial dimensions. This allows OibarNet to work with large input resolutions without needing to increase the network depth, and to potentially train it on different datasets without needing to alter the architecture. The \acrconnect{cnn}{-processed} features are then subject to a \gls{fc} layer back-end which estimates the pose via regression.

As part of the project, the temporal modelling of the \gls{cnn} features is also considered. This is investigated via the inclusion of a \gls{rnn} between the \gls{gap} output and the \gls{fc} input.

\subsubsection{Relative Pose Representation}

The first \gls{fc} block head maps the \gls{cnn} output to a \num{3}-vector estimate of the position, whereas the second head maps it to the \num{6}-dimensional estimate of the attitude formulated by \citet{zhou2018continuity}. The latter warrants special attention, as the \num{4}-dimensional quaternion, $\vq$, is normally used to represent the attitude of a spacecraft due to its low dimensionality and lack of singularities. However, its antipodal ambiguity-induced discontinuities (i.e., $\vq=-\vq$) have been shown to yield sub-optimal results in a deep learning environment compared to the 6D representation. The transformation $\vr\mapsto\mR$ from the 6D attitude representation to the direction cosine matrix representation involves reshaping $\vr$ into a $3\times 2$ matrix followed by a Gram-Schmidt orthogonalisation; the inverse transform $\mR\mapsto\vr$ simply involves discarding the rightmost column of $\mR$. Further details are given in Ref. \cite{zhou2018continuity}. For post-processing or error quantification, $\vq$ can then be obtained from $\mR$ using well-known isomorphisms \cite{markley2014fundamentals}.

\subsubsection{Loss Function}

The combination of predicted position and attitude quantities in a single loss function requires the incorporation of a scaling factor since these two quantities normally deal in different magnitudes \citep{proencca2020deep}. Typically, this scaling factor has been considered a hyperparameter part of the \gls{dnn}'s tuning process, which is sub-optimal.

In contrast, OibarNet follows the approach of \citet{cipolla2018multitask} and attributes one weight each to the position and attitude, $\sigma_\urt$ and $\sigma_\urr$, respectively, which become learnables and converge during the training process. The weights represent the task-specific variances of two Gaussian distributions, yielding a combined $\normltwo$ norm loss:

\begin{equation}
    \gL = \gL_{\urr}  \exp\left(-2 \hat{\sigma}_{\urr}\right) + \gL_\urt \exp\left(-2 \hat{\sigma}_\urt\right) + 2\left(\hat{\sigma}_{\urr} + \hat{\sigma}_\urt \right), \label{eq:method-losss2}
\end{equation}

\noindent where

\begin{equation}
    \gL_{\urr} = \sum\limits_{i=1}^B \lVert \hat{\vr}^{(i)} - \vr^{(i)} \rVert, \   \gL_{\urt} = \sum\limits_{i=1}^B \lVert \hat{\vt}^{(i)} - \vt^{(i)} \rVert.
\end{equation}

\noindent Here, $\hat{\vr},\hat{\vt}$  are the 6D attitude and 3D position estimated by the network, respectively; $\vr,\vt$ are the corresponding ground truths; $\lVert \cdot \rVert$ denotes the $\normltwo$ norm; and $B$ is the batch size.

\section{Demonstration and Testing}
\label{sec:demtest}

\noindent In this section, the methodology adopted for demonstrating the reliability of the developed \acrconnect{ai}{-based} solution for space docking and refuelling under \gls{oibar}. The validation tests are divided into three fronts: hardware validation, software validation, and integration validation. The former has been reported by \citet{lei2023novel}, the latter two are introduced herein.

\subsection{Software Validation}
\label{sec:dem;subsec:sw}

\noindent Supervised \gls{ml} algorithms require labelled and well-structured datasets not only for evaluation, but also for training. This is especially true, and even more relevant, for \acrconnect{dl}{-based} methods, which require large and diverse batches of data to learn how to generalise towards unseen scenarios due to the very large number of parameters at play.

However, labelled datasets for spacecraft pose estimation are scarce and expensive to obtain due to the intricate environmental conditions that must be emulated. As such, the first step in the software validation campaign for \gls{oibar} is to create a framework that allows for the generation of synthetic data: in contrast to real world sets, synthetic data is generally inexpensive and allows the possibility of having virtually unlimited samples.

Once this simulation environment is defined, the next step entails using it to produce synthetic docking trajectories. Lastly, an architecture selection round is performed based on the estimation performance on the data.

\subsubsection{Simulation Environment}
\label{sec:dem;subsec:sw;subsubsec:sim}

\noindent The simulation environment for \gls{oibar} is composed of two main components: a simulator designed in MATLAB/Simulink to replicate the orbital motion of the \gls{sv} and \gls{tv} under the influence of Earth; and an interface with the open-source 3D modelling software Blender\footnote{\url{https://www.blender.org}.} for the purpose of generating synthetic but realistic imagery of the \gls{tv} as viewed from the \gls{sv} on-board \gls{vbs} based on the states computed by the simulator.

The MATLAB/Simulink orbital simulator propagates \num{6}-\gls{dof} pose of a body orbiting Earth based on an initial state at a given date, which is obtained from \gls{tle} sets. The $J_2$ Earth oblateness and atmospheric drag perturbations are implemented for both acting forces and torques. Using the planetary ephemerides blocks available in Simulink, the relative states of Earth and the Sun are also computed. The SV motion trajectories are defined manually with respect to the target. 

\begin{figure*}[t]
  	\centering
  	\begin{subfigure}[t]{0.49\textwidth}\centering
  		\includegraphics[width=\textwidth]{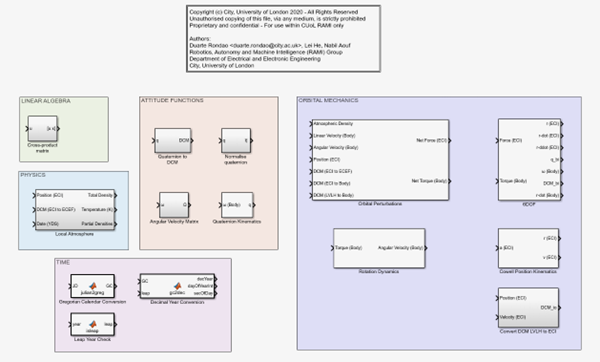}
      		\caption{\texttt{libOrbSim} modular library}
      	    \label{fig:demtest-orbsim-library}
  	\end{subfigure}
  	\begin{subfigure}[t]{0.49\textwidth}\centering
  		\includegraphics[width=\textwidth]{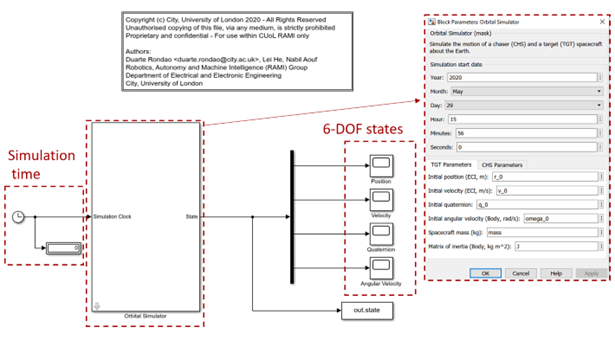}
      		\caption{High-level view of orbital simulator}
      	    \label{fig:demtest-orbsim-high}
  	\end{subfigure}
  \caption{MATLAB/Simulink orbital trajectory simulator environment.}
  \label{fig:demtest-orbsim}
\end{figure*}

\Figref{fig:demtest-orbsim} illustrates the developed orbital trajectory simulator environment. The approach to building the simulator followed a modular design (\figref{fig:demtest-orbsim-library}) which aimed to create basic blocks with fundamental functions, such as linear algebra, attitude manipulation and kinematics, and orbital mechanics, in order to facilitate any needed changes or customisation to the environment. The front-end (\figref{fig:demtest-orbsim-high}) allows for a simple configuration of initial states, including the date and \gls{tv} pose, permitting configurations for different scenarios.

\begin{figure*}[t]
  	\centering
  	\begin{subfigure}[t]{0.49\textwidth}\centering
  		\includegraphics[width=\textwidth]{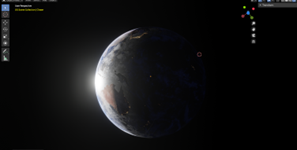}
      		\caption{Zoomed out view}
  	\end{subfigure}
  	\begin{subfigure}[t]{0.49\textwidth}\centering
  		\includegraphics[trim=0 5 0 0,clip,width=\textwidth]{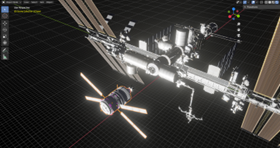}
      		\caption{Detailed view}
  	\end{subfigure}
  \caption{Blender 3D rendering environment.}
  \label{fig:demtest-blender}
\end{figure*}

The propagated states are then saved and interfaced with Blender, allowing the recreation of realistic images of the \gls{rvdb} dynamics of an Earth-orbiting \gls{sv} and \gls{tv} to be used for algorithm testing. \Figref{fig:demtest-blender} illustrates an example of a simulated scene within Blender. Besides the \gls{tv} and \gls{sv}, the Sun and Earth states have both been imported as well, ensuring a correct correspondence to the simulated time of day and solar phase angles.

\subsubsection{Synthetic Dataset Generation}
\label{sec:dem;subsec:sw;subsubsec:syndatagen}

Using the simulator presented in Subsection \ref{sec:dem;subsec:sw;subsubsec:sim}, a synthetic \gls{vbs} docking dataset was generated to validate the \acrconnect{ai}{-based} navigation system. The simulated scenario was chosen to be a refuelling of the \gls{iss} by the Kepler \gls{atv}. The \gls{cad} model of the \gls{oibar} docking mechanism was imported into the Blender environment and attached to the vehicles: the end-effector replaced the \gls{atv}’s existing \gls{rvdb} system, and the berthing fixture replaced the docking ports on the \gls{iss}. Due to a size mismatch, the original \gls{oibar} \glspl{cad} were scaled up to accommodate the vehicles within the simulation. In total, six different docking port locations on the \gls{iss} were considered; the inclusion of several docking ports was deemed beneficial as it grants diversity in terms of backgrounds, approach vectors (i.e., R-bar and V-bar) and illumination conditions.

The MATLAB/Simulink simulator are used to create the trajectory of the \gls{iss}, as well as the Earth and Sun states. The \gls{atv} \gls{sv} guidance trajectories are generated directly relative to the \gls{tv} frame; in particular, relative to the docking port considered for docking. Each \gls{sv} trajectory begins at a relative shaft-plane-berthing-plane distance of \SI{10}{\metre} and consists of three parts:

\begin{enumerate}[label={\arabic*)}]

\item \textbf{Acquisition.} This stage is characterised by a large cross-track motion ($x-y$ plane) whereby the \gls{vbs} is acquiring the target, prior to the end-effector and berthing fixture axes coinciding, but keeping the relative attitudes aligned. The \gls{sv} translates between two randomly generated waypoints at radial distances between \SIrange{1}{2}{\metre} from the alignment axis, before reducing this distance to zero. The linear velocity in this stage varies from \SIrange{0.09}{0.12}{\metre\per\second}.

\item \textbf{Forced translation.} Once the previous stage is concluded, the \gls{sv} end-effector and \gls{rv} berthing fixture are aligned in terms of a common along-track axis ($z$-axis), still at a relative distance of \SI{10}{\metre}. The \gls{sv} then translates along this axis at a nominal velocity of \SI{0.03}{\metre\per\second} to close the distance until \SI{3}{\metre}. To add variations amongst sequences, small perturbations are randomly generated and added to the translational and rotational motions; this is achieved by modelling a simple \gls{pi} controller and generating the next pose state from the feedback error. The magnitudes of the allowed perturbations vary from \SI{\pm 0.002}{\metre\per\second} for the along-track velocity, \SI{\pm0.01}{\metre} for the cross-track position, and \SI{\pm 0.1}{\degree} for the attitude; all with a probability of occurrence of \SI{10}{\percent}.

\item \textbf{Alignment and soft-docking.} This final stage begins with the position- and attitude-wise alignment of the fluidic and berthing planes from whatever misalignment state the previous stage may have ended in. The \gls{sv} then translates towards the \gls{tv}, while keeping an aligned attitude, until the spring protruding pin enters the central cavity, the latches are engaged, and soft-docking is achieved.

\end{enumerate}

\begin{figure}[t]
  	\centering
  	\begin{subfigure}[t]{\columnwidth}\centering
  		\includegraphics[width=0.32\textwidth]{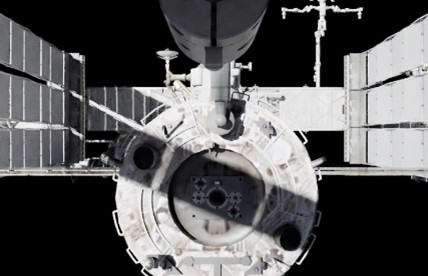}%
  	    \hfill
  		\includegraphics[width=0.32\textwidth]{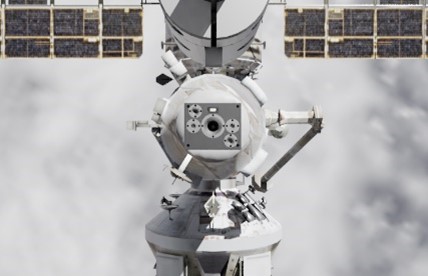}%
  	    \hfill
  		\includegraphics[width=0.32\textwidth]{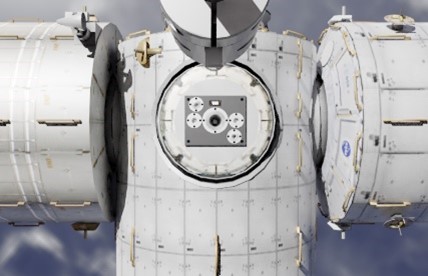}
  		\caption{\texttt{synthetic/iss} samples}
  	\end{subfigure}\\
  	\begin{subfigure}[t]{\columnwidth}\centering
  		\includegraphics[width=0.32\textwidth]{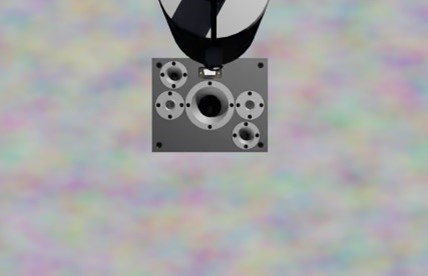}%
  	    \hfill
  		\includegraphics[width=0.32\textwidth]{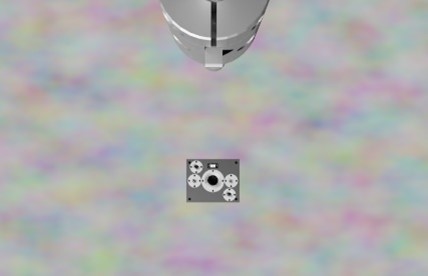}%
  	    \hfill
  		\includegraphics[width=0.32\textwidth]{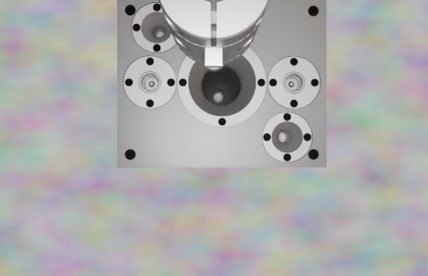}
  		\caption{\texttt{synthetic/perlin} samples}
  	\end{subfigure}
  \caption{Sample frames from \acrshort{oibar}’s synthetic dataset.}
  \label{fig:demtest-syntheticsamples}
\end{figure}

The average sequence duration is $\sim$\SI{5}{\minute}, and the synthetic dataset comprises \num{12} sequences in total. Furthermore, the synthetic dataset is composed of two subsets. The first one, \texttt{synthetic/iss}, consists of nominal scenario conditions where the docking mechanism is mounted on one of the docking ports of the \gls{iss}. The second subset, \texttt{synthetic/perlin}, models similar relative trajectories, but removes all meshes except for the docking mechanism, replacing the background with randomised Perlin noise. The objective of \texttt{synthetic/perlin} is to complement the \texttt{synthetic/iss} subset to provide additional training data and to help OibarNet focus on extracting features of the target and ignore the background and environment. \Tabref{tab:dem-syntheticdata0} summarises the characteristics of the generated dataset. The Sun elevation angles were selected according to daylight conditions for each specific docking port, whereby some variance was introduced by either selecting sunrise or sunset periods. \Figref{fig:demtest-syntheticsamples} illustrates a few sample frames from the dataset.

\begin{table}[t]
\sisetup{detect-weight=true,detect-inline-weight=math,separate-uncertainty}
\centering
 \caption{Synthetic dataset characteristics used for training, validation, and testing of OibarNet.}
 \label{tab:dem-syntheticdata0}
  \begin{adjustbox}{width={\columnwidth},totalheight={\textheight},keepaspectratio}
 \begin{tabular}{l c c c S c c} 
	\toprule
	\multirow{2}{*}{\textbf{Sequence}}  & \multirow{2}{*}{\textbf{Docking port}} & \multicolumn{2}{c}{\textbf{Approach axis}}  & {\multirow{2}{*}{\textbf{Sun elevation angle (deg)}}} & \multicolumn{2}{c}{\textbf{Background}}\\
                                    	&                                        & {\textbf{V-bar}}	& {\textbf{R-bar}}  &                                                     & {\textbf{\texttt{iss}}} & {\textbf{\texttt{perlin}}}\\
	\midrule
	1   &   1   &   $+$ &       &   37  &   $\times$    &           \\
	2   &   1   &   $+$ &       &   75  &               &  $\times$ \\
	3   &   2   &       & $-$   &   56  &   $\times$    &           \\
	4   &   2   &       & $-$   &   146 &               & $\times$  \\
	5   &   3   &   $-$ &       &   127 &               & $\times$  \\
	6   &   3   &   $-$ &       &   165 &   $\times$    &           \\
	7   &   4   &       & $-$   &   56  &               & $\times$  \\
	8   &   4   &       & $-$   &   146 &  $\times$     &           \\
	9   &   5   &       & $+$   &   56  &  $\times$     &           \\
	10  &   5   &       & $+$   &   146 &               & $\times$  \\
	11  &   6   &       & $+$   &   56  &               & $\times$  \\
	12  &   6   &       & $+$   &   146 &  $\times$     &           \\
	\bottomrule
 \end{tabular}
 \end{adjustbox}
\end{table}

The synthetic dataset emulates the \gls{vbs} used in the integration testing (see Section 4.4, Table 3); images are generated at a resolution of \qtyproduct[product-units = single]{744 x 480}{\pixel} and a framerate of \SI{10}{\hertz}.

\subsubsection{Training}
\label{sec:dem;subsec:sw;subsubsec:training}

From \Tabref{tab:dem-syntheticdata0}, sequences 1 and 8 were selected exclusively for testing, whereas the remaining sequences were used for training.

Further, the latter were divided according to a \SI{80}{\percent}/\SI{20}{\percent} partition to include validation data as well and allow the unbiased benchmarking of different models. Since the data consists of image sequences, these were partitioned into smaller ones to accomplish this, where the length was randomly sampled from a range of powers of two, resulting in an interval of $\{\num{64},\ldots,\num{1024}\}$ seconds.

Image augmentation was performed online on the training data to prevent overfitting. This was applied in two fronts. The first one is related to transform operations in the \gls{ip} domain: randomised changes in terms of brightness, contrast, colour, Gaussian noise and blur, for example, are generated to robustify the network against potentially unpredicted imaging conditions during deployment. The second one is related to operations in the pose domain, whereby randomly generated perspective transforms are applied to images in the sequence to simulate deviations in the trajectory (i.e., translation shifts, in-plane rotations, homography-induced off-plane rotations). The latter is of particular importance since, despite the sequence partitioning for training, the forced translation phase dominates each sequence, generating an imbalance on the distribution of position states.

OibarNet is implemented in MATLAB R2021b using a custom-developed library. Models are trained for \num{100} epochs with a cyclical learning rate decay of \num{5} cycles \citep{smith2017cyclical}. The Adam optimiser \cite{kingma2014adam} is used. A dropout probability of \num{0.2} is used. Training is performed on City, University of London’s high performance computing facility Hyperion using one NVIDIA® Quadro RTX™ 8000 \gls{gpu} with \SI{48}{\giga\byte} VRAM.

\subsubsection{Training}

The test results are presented in terms of the position and attitude error metrics, respectively:

\begin{align}
\delta\tilde{t} &\coloneqq \lVert \hat{\vt} - \vt \rVert,\\
\delta\tilde{q} &\coloneqq 2\arccos\left(\hat{\vq}^{-1} \otimes \vq\right)_4,
\end{align}

\noindent where the subscript ``\num{4}'' denotes the scalar element of the resulting quaternion. Additionally, the position error is also assessed in terms of the relative range:

\begin{equation}
\delta\tilde{t}_\urr \coloneqq \dfrac{\delta\tilde{t}}{\lVert \vt \rVert}.
\end{equation}

\subsubsection{CNN Architecture Selection}

From the relative pose estimation point of view, a few differences may be expected between an RV manoeuvre and a docking sequence. Firstly, a reduced variation in the attitude is expected during docking since the \gls{sv} is expected to be inside the cone-shaped approach corridor of the \gls{tv} \citep{fehse2003sensors}; in opposition, the target may be tumbling during \gls{rv}. Secondly, an increased apparent variation in the position can be expected during docking, as due to the reduced relative distance any small shift will result in a large displacement of the \gls{tv} berthing fixture in the \gls{fov}.

To better assess the influence of these factors, multiple \gls{cnn} architectures are benchmarked. The baseline is Darknet-19 \cite{redmon2017yolo9000}, which has successfully been applied in the past to the problem of pose estimation in RV \cite{rondao2021chinet}. To analyse the effect of increasing the capacity of the model, Darknet-53 \cite{redmon2018yolov3} and ResNet-101 \cite{he2015deep} are included. Lastly, it is also important to verify the change in performance when reducing the capacity, and SqueezeNet \cite{iandola2016squeezenet} is thus included in the benchmark as well. 

\begin{figure}[t]
  \centering
  \includegraphics[width=\columnwidth]{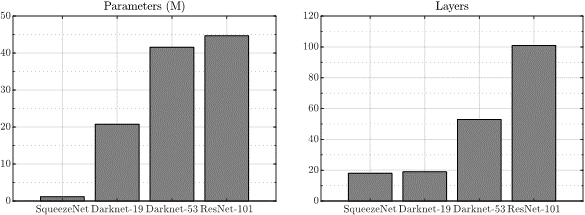}
  \caption{Characteristics of the four different benchmarked CNN models.}
  \label{fig:demtest-cnnmodels}
\end{figure}

\Figref{fig:demtest-cnnmodels} summarises the number of parameters, in millions, and number of layers of the four different \gls{cnn} models considered for benchmarking.

\subsection{Integration Validation Methodology}

\noindent The goal of the integration testing is to validate the combination of the hardware and software blocks outlined above. In this setup, the navigation \gls{vbs} is incorporated into the hardware setup \citep{lei2023novel} to acquire a stream of images to be processed by the navigation algorithm (\secref{sec:dem;subsec:sw}) during the docking manoeuvre emulated by the robotic setup.

\begin{figure}[t]
  	\centering
  	\begin{subfigure}[t]{0.59\columnwidth}\centering
  		\includegraphics[height=7\baselineskip]{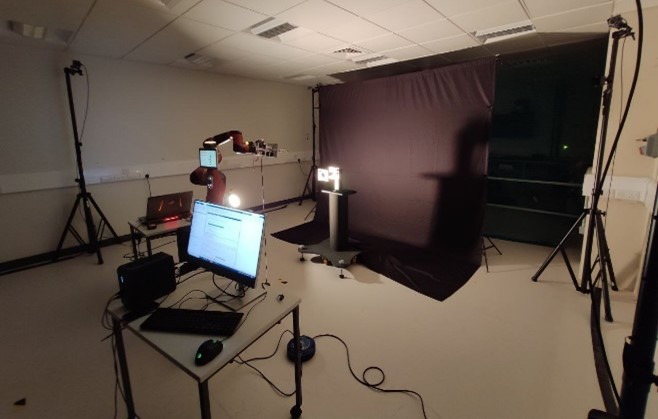}
      		\caption{Laboratory setup}
  	\end{subfigure}\hfill
  	\begin{subfigure}[t]{0.39\columnwidth}\centering
  		\includegraphics[height=7\baselineskip]{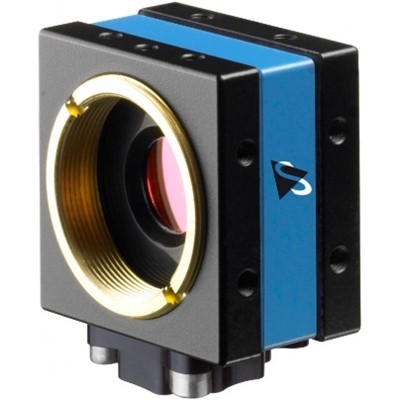}
      		\caption{DFK 22BUC03 \gls{vbs}}
  	\end{subfigure}
  \caption{Integration validation setup at City, University of London’s \acrshort{asmil}.}
  \label{fig:demtest-intsetup}
\end{figure}

 \Figref{fig:demtest-intsetup} illustrates the integration validation setup. A blackout backdrop is placed behind the target berthing fixture to simulate the imaging conditions of a featureless deep space background. The target itself is illuminated by a single \SI{400}{\watt} directional floodlight. The \gls{sv} and \gls{tv} are placed inside the capture volume of an OptiTrack\footnote{\url{https://optitrack.com}.} motion capture system for recording the ground truth measuring approximately \qtyproduct[product-units = single]{5 x 5 x 3}{\metre}. OptiTrack can record 6-\gls{dof} pose data of rigid and flexible bodies by detecting, tracking, and triangulating passive near infrared markers placed on targets. The data can be saved or stream over a local network in real-time. 

The OptiTrack setup at City consists in six PrimeX 13 cameras with a resolution of \qtyproduct[product-units = single]{1280 x 1024}{\pixel} running at a native framerate of \SI{240}{\hertz}, capable of achieving positional errors less than \SI{\pm 0.20}{\milli\metre} and rotational errors less than \SI{0.5}{\degree}.

The used \gls{vbs} is the Imaging Source DFK 22BUC03 colour camera with a \SI[parse-numbers = false]{\sfrac{1}{3}}{inch} format CMOS sensor (Onsemi MT9V024) and a native resolution of \qtyproduct[product-units = single]{744 x 480}{\pixel}, fitted with a Kowa LM4NCL \SI{3.5}{\milli\metre} focal length lens. \Tabref{tab:dem-vbs} summarises the technical characteristics of the \gls{vbs}.

\begin{table}[t]
\sisetup{detect-weight=true,detect-inline-weight=math,separate-uncertainty}
\centering
 \caption{Technical data -- DFK 22BUC03.}
 \label{tab:dem-vbs}
 \begin{tabular}{l c S} 
	\toprule
	\multicolumn{1}{c}{\textbf{Parameter}} & \textbf{Units} & {\textbf{Value}}\\
	\midrule
	Resolution   &   \SI{}{\pixel}   &   \numproduct{744x480} \\
	Maximum frame rate & \SI{}{\hertz} & \num{76}\\
	Focal length    & \SI{}{\milli\metre} & \num{3.5}\\
	Horizontal \acrshort{fov} & \SI{}{\degree} & \num{65.6}\\
	Vertical \acrshort{fov} & \SI{}{\degree} & \num{44.7}\\
	\bottomrule
 \end{tabular}
\end{table}

The workstation consists of an Intel® NUC \num{9} Pro with an NVIDIA® RTX™ \num{3060} Ti Mini GPU with \SI{8}{\giga\byte} VRAM. The workstation is used for both experimental data offline validation of OibarNet and real-time online testing of the network, at a framerate of \SI{10}{\hertz}.

\subsubsection{Experimental Dataset Generation}

The docking imaging sequences acquired with the experimental setup follow the same structure as the synthetic dataset (\secref{sec:dem;subsec:sw;subsubsec:syndatagen}) albeit with two key differences. The first one is that all experimental sequences feature the same type of background (black, deep space). The second is that, rather than implementing \acrconnect{pi}{induced} pose perturbations during the forced translation (Phase 2), a static misalignment of the pose is randomly introduced in each sequence at the beginning of the phase, which is then corrected at the beginning of the final one.

In total, \num{12} experimental trajectories are collected, whereby the angle of illumination alternates between port and starboard. The average sequence duration is $\sim$\SI{3.15}{\minute}. The first \num{10} sequences are used for training and validation of the model according to the methodology of \Secref{sec:dem;subsec:sw;subsubsec:training}. Sequences \texttt{experimental/11} and \texttt{experimental/12} are used exclusively for testing.

\subsubsection{Ground Truth Calibration Toolbox}

The OptiTrack system used to record the ground truth measures the poses of rigid bodies equipped with infrared markers. However, it does not directly output the relative pose between the \gls{vbs} and the \gls{tv} (as illustrated in \figref{fig:relwork-bodyframes}, \secref{sec:relwork}), which is required by the navigation algorithm.

\begin{figure}[t]
  \centering
  \includegraphics[width=\columnwidth]{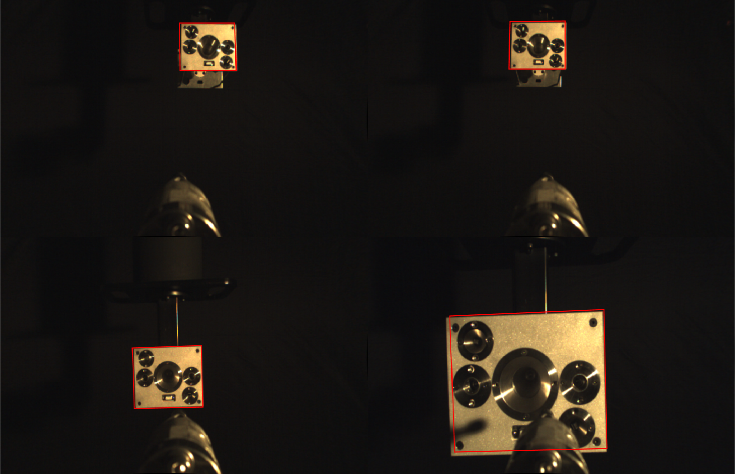}
  \caption{Ground truth calibration toolbox output, visualised on some frames of the experimental dataset by reprojecting the target's \gls{cad} model (in red) according to the measured pose.}
  \label{fig:demtest-calibresult}
\end{figure}

To this end, a toolbox was developed in MATLAB to calibrate the output OptiTrack data and generate the required relative pose, based on the work of \citet{valmorbida2020calibration}. The output of the calibration toolbox are the static transforms $\mT_{ic}$, mapping the camera frame $\Fc$ to the frame of reference $\Fi$ defined by the physical markers placed on its housing and tracked by OptiTrack, and $\mT_{sb}$, mapping the target's body frame $\Fb$ to the frame of reference $\Fs$ defined by the markers placed on it. These transforms then make possible to map the OptiTrack marker-defined rigid bodies' poses, which are measured relative to $\Fo$, the system's arbitrary global frame of reference, into a usable ground truth $\mT_{bc}$ defined in terms of the \gls{vbs} frame of reference. \Figref{fig:demtest-calibresult} illustrates the output of the calibration procedure on select samples of the dataset.

\section{Results}

\subsection{Software Testing}

\subsubsection{\acrshort{cnn} Model Benchmarking}

\begin{figure}[t]
  \centering
  \includegraphics[width=\columnwidth]{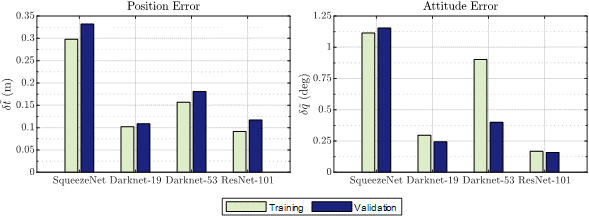}
  \caption{Average training and validation pose estimation errors for different \acrshort{cnn} architectures trained on the \texttt{synthetic} dataset.}
  \label{fig:res-cnnbenchmark}
\end{figure}

\Figref{fig:res-cnnbenchmark} illustrates the results of the different models trained on the \texttt{synthetic} dataset, presented in terms of mean position and attitude errors averaged per trajectory. It can be seen that the performance of SqueezeNet is considerably worse than the baseline Darknet-19, yielding errors twice as large for both position and attitude. Both networks have a very similar number of layers, but Darknet-19 has substantially more learnable parameters (as indicated in \figref{fig:demtest-cnnmodels}); the reduced attitude variance in the docking manoeuvres is thus shown not to justify a decrease in parameters.

Interestingly, the error for Darknet-53 actually increases with respect to the baseline. Once the capacity of the CNN is further increased with ResNet-101, though, the error decreases again, making the network the best performing model (except on position validation error, which is slightly larger than Darknet-19’s).

\begin{figure}[t]
  	\centering
  	\begin{subfigure}[t]{0.49\columnwidth}\centering
  		\includegraphics[width=\textwidth]{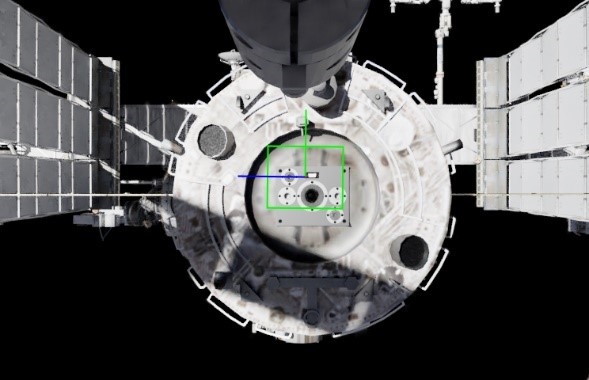}
  		\caption{SqueezeNet}
    \end{subfigure}%
    \hfill
  	\begin{subfigure}[t]{0.49\columnwidth}\centering
  		\includegraphics[width=\textwidth]{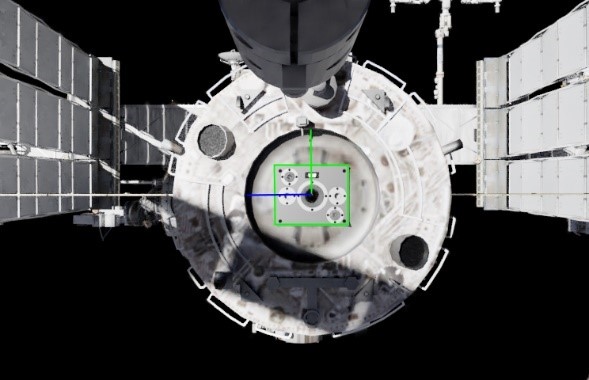}
  		\caption{Darknet-19}
    \end{subfigure}\\
  	\begin{subfigure}[t]{0.49\columnwidth}\centering
  		\includegraphics[width=\textwidth]{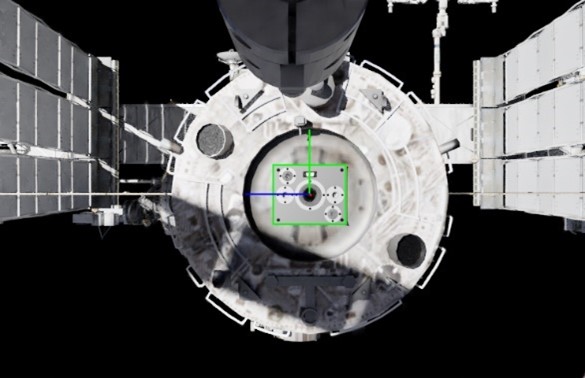}
  		\caption{Darknet-53}
    \end{subfigure}%
    \hfill
  	\begin{subfigure}[t]{0.49\columnwidth}\centering
  		\includegraphics[width=\textwidth]{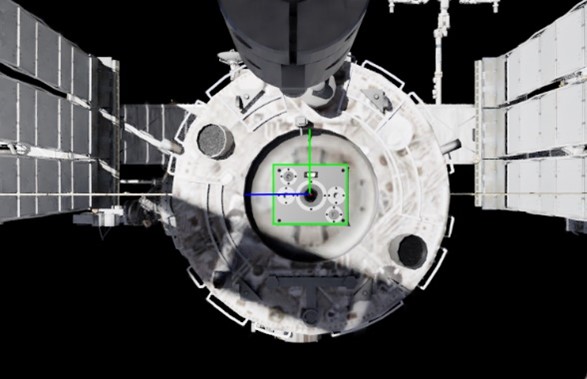}
  		\caption{ResNet-101}
    \end{subfigure}\\
  \caption{Qualitative pose estimation performance on a validation sequence of the \texttt{synthetic} dataset for different \acrshort{cnn} models.}
  \label{fig:res-qualitative-synthetic-valid}
\end{figure}

The results of \Figref{fig:res-cnnbenchmark} are presented with the caveat that they represent average errors per trajectory, but where the data is not composed of random images but time sequences. As such, while a histogram visualisation is useful for a first analysis of each model’s performance, it is also important to look at how these perform in specific, individual situations. For example, \Figref{fig:res-qualitative-synthetic-valid} represents the qualitative performance of each model on a single frame of one of the synthetic validation sequences; the rectangular boundary of the berthing fixture is reprojected in green using the predicted pose, and the axes of the estimated frame $\Ft$ are also shown. The results show that SqueezeNet is overfitting at least on the position state, as it expects the berthing fixture to be located in the centre of the \gls{fov}, when in reality the \gls{sv} end-effector is still misaligned. The other three models with increased capacity demonstrate no issues in estimating the correct relative position.

\begin{figure}[t]
  	\centering
  	\begin{subfigure}[t]{0.49\columnwidth}\centering
  		\includegraphics[width=\textwidth]{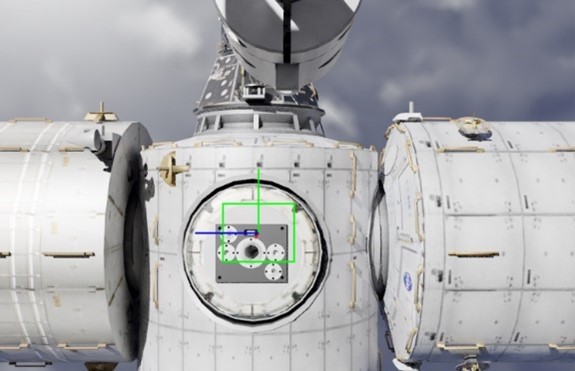}
  		\caption{SqueezeNet}
    \end{subfigure}%
    \hfill
  	\begin{subfigure}[t]{0.49\columnwidth}\centering
  		\includegraphics[width=\textwidth]{figs/results/benchmark-synthetic-training-dn19.jpg}
  		\caption{Darknet-19}
    \end{subfigure}\\
  	\begin{subfigure}[t]{0.49\columnwidth}\centering
  		\includegraphics[width=\textwidth]{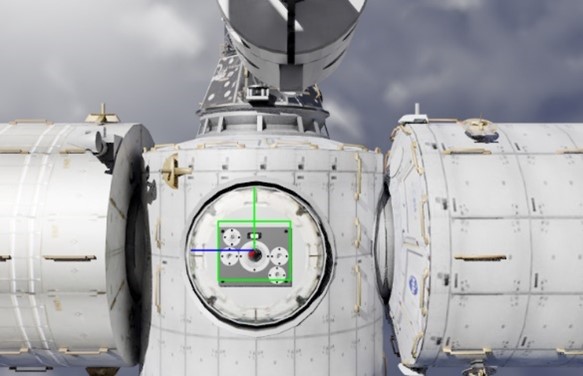}
  		\caption{Darknet-53}
    \end{subfigure}%
    \hfill
  	\begin{subfigure}[t]{0.49\columnwidth}\centering
  		\includegraphics[width=\textwidth]{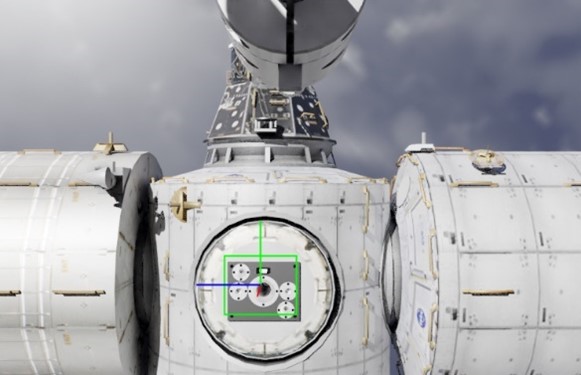}
  		\caption{ResNet-101}
    \end{subfigure}\\
  \caption{Qualitative pose estimation performance on a training sequence of the \texttt{synthetic} dataset for different \acrshort{cnn} models.}
  \label{fig:res-qualitative-synthetic-train}
\end{figure}

Consider now, however, the performance on one training sequence, as illustrated in   \Figref{fig:res-qualitative-synthetic-train}: SqueezeNet (a) is shown to be underfitting, but so is Darknet-19 (b). This suggests that increasing the capacity would benefit OibarNet, as confirmed by the frame output by Darknet-53 (c) showing a better fit, despite the summary metrics in \Figref{fig:res-qualitative-synthetic-valid}. The performance with ResNet-101 (d) is slightly better even, confirming it as the choice for the final \gls{cnn} model in OibarNet.

\subsubsection{Performance Evaluation on Synthetic Dataset}
\label{sec:res;subsec:software;subsubsec:synthetic}

In this subsection, the performance of the navigation algorithm is evaluated on the test sequences \texttt{synthetic/01} and \texttt{synthetic/08}, as outlined in \Secref{sec:demtest}, and according to the selected ResNet-101 \gls{cnn} architecture for OibarNet. 

\begin{figure}[t]
  	\centering
  	\begin{subfigure}[t]{0.49\columnwidth}\centering
  		\includegraphics[width=\textwidth]{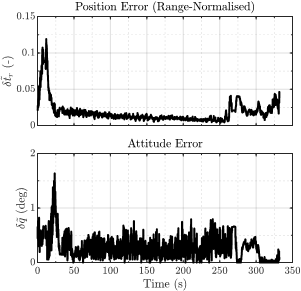}
  		\caption{\texttt{synthetic/01}}
    \end{subfigure}%
    \hfill
  	\begin{subfigure}[t]{0.49\columnwidth}\centering
  		\includegraphics[width=\textwidth]{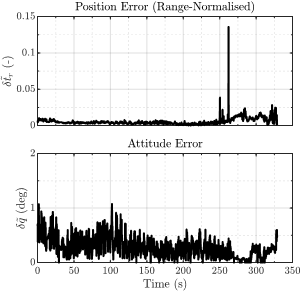}
  		\caption{\texttt{synthetic/08}}
    \end{subfigure}%
  \caption{Estimated position and attitude errors over time on the two test sequences of the synthetic dataset.}
  \label{fig:res-quantitative-synthetic}
\end{figure}

\Figref{fig:res-quantitative-synthetic} showcases the attained pose estimation errors for each sequence using the final OibarNet model; the position errors are normalised as a percentage of range. \Tabref{tab:res-quantitative-synthetic} summarises these statistics.

\begin{table}[t]
\sisetup{detect-weight=true,detect-inline-weight=math,separate-uncertainty}
\centering
 \caption{Summary performance statistics on the two test sequences of the synthetic dataset.}
  \label{tab:res-quantitative-synthetic}
  \begin{adjustbox}{width={\columnwidth},totalheight={\textheight},keepaspectratio}
 \begin{tabular}{l S S S S S S} 
	\toprule
	\multicolumn{1}{c}{\multirow{2}{*}{\textbf{Sequence}}} & \multicolumn{2}{c}{\textbf{Position Error (\si{\metre})}} & \multicolumn{2}{c}{\textbf{Attitude Error (\si{\degree})}} & \multicolumn{2}{c}{\textbf{Requirement Compliance (\si{\percent})}}\\
	\cmidrule(lr){2-3}
	\cmidrule(lr){4-5}
	\cmidrule(lr){6-7}
	& \multicolumn{1}{c}{\textbf{Mean}} & \multicolumn{1}{c}{\textbf{Median}} & \multicolumn{1}{c}{\textbf{Mean}} & \multicolumn{1}{c}{\textbf{Median}} & \multicolumn{1}{c}{\textbf{Position}} & \multicolumn{1}{c}{\textbf{Attitude}}\\
	\midrule
	\texttt{synthetic/01} & 1.81 & 1.39 & 0.29 & 0.26 & 96.33 & 100\\
	\texttt{synthetic/08} & 0.56 & 0.43 & 0.28 & 0.26 & 99.91 & 100\\
	\bottomrule
 \end{tabular}
 \end{adjustbox}
\end{table}

The figures demonstrate that, for \texttt{synthetic/01}, OibarNet fulfils the \SI{5}{\percent} maximum range-normalised position error requirement (defined in Ref.~\cite{lei2023novel}) for most of the trajectory. The exception is a segment corresponding to phase 1 (acquisition) whereby the \gls{sv} moves to a waypoint representing a large displacement relative to the alignment axis, representing about \SI{3.7}{\percent} of the sequence’s duration. After this period, the error converges to values below \SI{2.5}{\percent} of range, further decreasing as the \gls{sv} closes in on the \gls{tv}, until the beginning of phase 3 (alignment and soft-docking), where the very short range causes the error to rise, but not above the requirement threshold. The attitude estimation performance is shown to fully comply with the \SI{5}{\degree} maximum error requirement.

The position estimation performance of the navigation algorithm on \texttt{synthetic/08} is observed to be better than the previous sequence, as an improvement of \num{1.25} percent points on the mean value and \num{0.96} percent points on the median value are achieved. Furthermore, the position estimate is virtually fully compliant with the defined requirement, save for a singular spike (less than \SI{0.1}{\percent} of the trajectory). The attitude estimation is again entirely compliant and practically does not surpass \SI{1}{\degree} in error.

\begin{figure}[t]
  	\centering
  	\begin{subfigure}[t]{0.32\columnwidth}\centering
  		\includegraphics[width=\textwidth]{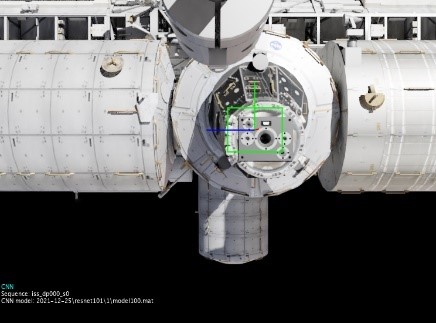}%
  		\caption{$\tau=\SI{20}{\second}$}
  		  	\end{subfigure}
  	    \hfill
  	      	\begin{subfigure}[t]{0.32\columnwidth}\centering
  		\includegraphics[width=\textwidth]{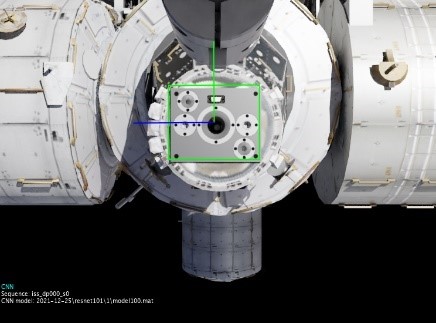}%
  		\caption{$\tau=\SI{160}{\second}$}
  		  	\end{subfigure}
  	    \hfill
  	      	\begin{subfigure}[t]{0.32\columnwidth}\centering
  		\includegraphics[width=\textwidth]{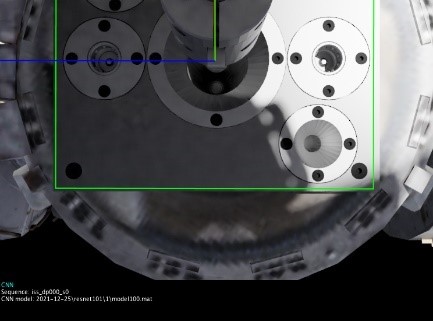}
  		\caption{$\tau=\SI{305}{\second}$}
  	\end{subfigure}
  	  \caption{Qualitative pose estimation performance on the \texttt{synthetic/01} test sequence.}
  \label{fig:res-qualitative-synthetic-01}
\end{figure}

\begin{figure}[t]
  	\centering
  	\begin{subfigure}[t]{0.32\columnwidth}\centering
  		\includegraphics[width=\textwidth]{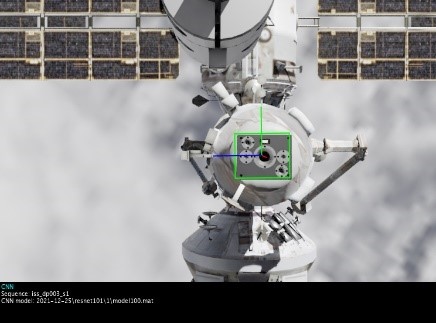}%
  		\caption{$\tau=\SI{20}{\second}$}
  		  	\end{subfigure}
  	    \hfill
  	      	\begin{subfigure}[t]{0.32\columnwidth}\centering
  		\includegraphics[width=\textwidth]{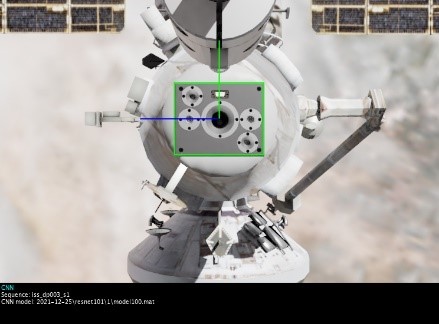}%
  		\caption{$\tau=\SI{155}{\second}$}
  		  	\end{subfigure}
  	    \hfill
  	      	\begin{subfigure}[t]{0.32\columnwidth}\centering
  		\includegraphics[width=\textwidth]{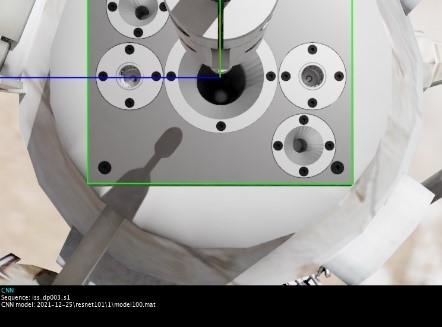}
  		\caption{$\tau=\SI{289}{\second}$}
  	\end{subfigure}
  	  \caption{Qualitative pose estimation performance on the \texttt{synthetic/08} test sequence.}
  \label{fig:res-qualitative-synthetic-08}
\end{figure}

 \Figref{fig:res-qualitative-synthetic-01} and \Figref{fig:res-qualitative-synthetic-08} exhibit some frames from each sequence with the respective qualitative pose estimation fit overlaid. On \texttt{synthetic/01} the \gls{tv} structure surrounding the berthing fixture is quite complex, which from the \gls{ip} point of view represents a more challenging background than the case of \texttt{synthetic/08}, despite it being an R-bar trajectory which includes Earth. Additionally, the illumination conditions on the former appear to make the berthing fixture harder to distinguish from the \gls{iss} structure relative to the latter. Both aspects could provide an explanation to the increased position error seen in the beginning of \texttt{synthetic/01}.

\subsubsection{Effect of Temporal Modelling}

The designed OibarNet pipeline uses a \gls{cnn} front-end to process incoming images and extract features. However, these images are processed individually, whereby the data as a whole represents a time sequence depicting a docking manoeuvre, which implies that each sample is time correlated. Modifications to \gls{cnn} architectures have been proposed in the past to account for this correlation and shown to improve the relative pose estimation error for rendezvous. Specifically, \glspl{drcnn} include a recurrent sub-network as the back-end of the pipeline that models the features extracted by the \gls{cnn} \cite{rondao2021chinet}.

This test investigates the effect of applying a \gls{drcnn} to the problem of relative pose estimation for docking. To this end, the trained \gls{cnn} model was appended with a recurrent model, further trained on the output of the \gls{cnn} for the same dataset, consisting of \gls{bilstm} cells \cite{graves2005bidirectional}. Contrary to regular \glspl{lstm}, \glspl{bilstm} run sequence inputs in two directions: one from past to future, and the other from future to past, thus preserving information from both past and future. This feature can be beneficial for \gls{rvdb} pose estimation problems since trajectories are continuous, meaning that not only do the previous states influence the present, but states in the future provide context to the preceding ones.

\begin{figure}[t]
  \centering
  \includegraphics[width=\columnwidth]{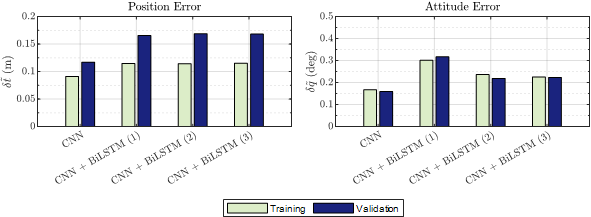}
  \caption{Effect of adding recurrent neural layers on the training and validation performance of the synthetic dataset. The numbers in parenthesis denote the number of recurrent layers.}
  \label{fig:res-pose-recurrent}
\end{figure}

The results of the benchmark are illustrated in \Figref{fig:res-pose-recurrent}. It can be seen that the addition of a recurrent layer degrades not only the validation performance, but also the training performance; this is witnessed both in terms of position and attitude estimation. Adding more recurrent layers lowers the error on the attitude estimate, but even with three layers this is still higher than that obtained for the CNN alone. Furthermore, the position error is shown not to decrease.

This study represents an interesting result since it is seemingly counter-intuitive and diverges from the findings reported by \citet{rondao2021chinet}. However, whereas the apparent relative motion during an \gls{rv} is typically smooth and predictable (e.g., \gls{sv} at a hold point observing the \gls{tv} tumbling), the docking trajectories modelled within the scope of \gls{oibar} are actually more dynamic and include higher stochasticity due to the random perturbations added during the approach phase. As such, one explanation towards the poor performance of the \gls{drcnn} in this case could be the failure in modelling these high-frequency, random changes in motion, in which the CNN indeed an advantage as it is processing each time-step individually.

Further avenues of research could still be pursued, however. For example, the inclusion of attention-based mechanisms remains to be investigated for \gls{rvdb}, where the network would be capable of self-learning weights to be attributed to each time-step in the sequence, thus becoming able to let certain segments influence the estimate more than others (e.g., placing less attention on the immediate perturbations and more on the overall along-track motion).

Due to the attained results, the OibarNet architecture was not altered for the integration tests.

\subsection{Integration Testing}

\subsubsection{Performance Evaluation on Experimental Dataset}
\label{sec:res;subsec:integration;subsubsec:experimental}

This section is analogous to \Subsecref{sec:res;subsec:software;subsubsec:synthetic} with the difference that the selected OibarNet \gls{cnn} architecture is evaluated and tested on experimental data collected in laboratory. As outlined in \Secref{sec:demtest}, the performance of the navigation algorithm is evaluated on the test sequences \texttt{experimental/11} and \texttt{experimental/12}. 

\begin{figure}[t]
  	\centering
  	\begin{subfigure}[t]{0.49\columnwidth}\centering
  		\includegraphics[width=\textwidth]{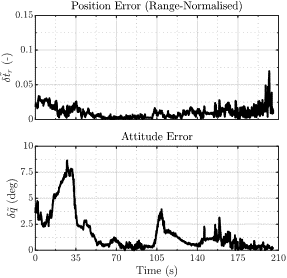}
  		\caption{\texttt{experimental/11}}
    \end{subfigure}%
    \hfill
  	\begin{subfigure}[t]{0.49\columnwidth}\centering
  		\includegraphics[width=\textwidth]{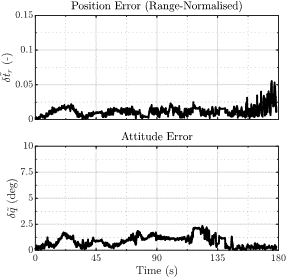}
  		\caption{\texttt{experimental/12}}
    \end{subfigure}%
  \caption{Estimated position and attitude errors over time on the two test sequences of the experimental dataset.}
  \label{fig:res-quantitative-experimental}
\end{figure}

\begin{figure}[t]
  	\centering
  	\begin{subfigure}[t]{0.32\columnwidth}\centering
  		\includegraphics[width=\textwidth]{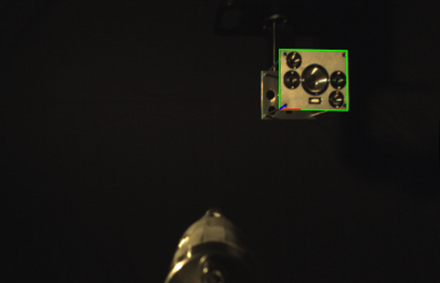}%
  		\caption{$\tau=\SI{20}{\second}$}
  		  	\end{subfigure}
  	    \hfill
  	      	\begin{subfigure}[t]{0.32\columnwidth}\centering
  		\includegraphics[width=\textwidth]{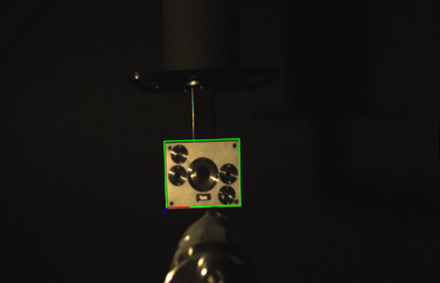}%
  		\caption{$\tau=\SI{160}{\second}$}
  		  	\end{subfigure}
  	    \hfill
  	      	\begin{subfigure}[t]{0.32\columnwidth}\centering
  		\includegraphics[width=\textwidth]{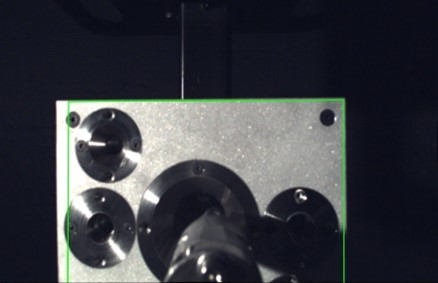}
  		\caption{$\tau=\SI{305}{\second}$}
  	\end{subfigure}
  	  \caption{Qualitative pose estimation performance on the \texttt{experimental/11} test sequence.}
  \label{fig:res-qualitative-experimental-11}
\end{figure}

\begin{figure}[t]
  	\centering
  	\begin{subfigure}[t]{0.32\columnwidth}\centering
  		\includegraphics[width=\textwidth]{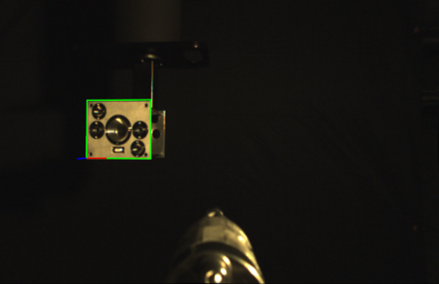}%
  		\caption{$\tau=\SI{20}{\second}$}
  		  	\end{subfigure}
  	    \hfill
  	      	\begin{subfigure}[t]{0.32\columnwidth}\centering
  		\includegraphics[width=\textwidth]{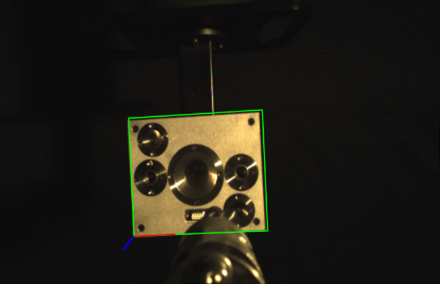}%
  		\caption{$\tau=\SI{160}{\second}$}
  		  	\end{subfigure}
  	    \hfill
  	      	\begin{subfigure}[t]{0.32\columnwidth}\centering
  		\includegraphics[width=\textwidth]{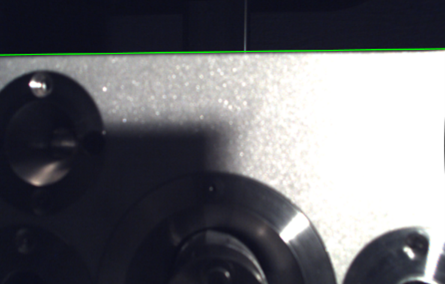}
  		\caption{$\tau=\SI{305}{\second}$}
  		 \label{fig:res-qualitative-experimental-12-3}
  	\end{subfigure}
  	  \caption{Qualitative pose estimation performance on the \texttt{experimental/12} test sequence.}
  \label{fig:res-qualitative-experimental-12}
\end{figure}

\Figref{fig:res-quantitative-experimental} displays the attained pose estimation errors for both trajectories. \Tabref{tab:res-quantitative-experimental} summarises the performance metric statistics. Lastly, \Figref{fig:res-qualitative-experimental-11} and \Figref{fig:res-qualitative-experimental-12} illustrate qualitative estimation results for a few frames of the \texttt{experimental/11} and \texttt{experimental/12} sequences, respectively. The relative position estimation error follows a similar trend to the synthetic dataset case: lower during the approach phase and increasing in the final alignment and soft-docking phase. Overall, the curves oscillate more in amplitude for both trajectories; this is a possible by-product of using real-data which can be contaminated with random errors (e.g., sensor noise) and systematic errors (e.g., errors in the motion capture system calibration), which are not scene in the ideal development conditions of synthetic datasets. The reduced number of training samples relative to the synthetic case also affects the solution (i.e., the experimental trajectories are shorter). Nevertheless, the requirement compliance is virtually \SI{100}{\percent} for both trajectories. 

\begin{table}[t]
\sisetup{detect-weight=true,detect-inline-weight=math,separate-uncertainty}
\centering
 \caption{Summary performance statistics on the two test sequences of the experimental dataset.}
  \label{tab:res-quantitative-experimental}
  \begin{adjustbox}{width={\columnwidth},totalheight={\textheight},keepaspectratio}
 \begin{tabular}{l S S S S S S} 
	\toprule
	\multicolumn{1}{c}{\multirow{2}{*}{\textbf{Sequence}}} & \multicolumn{2}{c}{\textbf{Position Error (\si{\metre})}} & \multicolumn{2}{c}{\textbf{Attitude Error (\si{\degree})}} & \multicolumn{2}{c}{\textbf{Requirement Compliance (\si{\percent})}}\\
	\cmidrule(lr){2-3}
	\cmidrule(lr){4-5}
	\cmidrule(lr){6-7}
	& \multicolumn{1}{c}{\textbf{Mean}} & \multicolumn{1}{c}{\textbf{Median}} & \multicolumn{1}{c}{\textbf{Mean}} & \multicolumn{1}{c}{\textbf{Median}} & \multicolumn{1}{c}{\textbf{Position}} & \multicolumn{1}{c}{\textbf{Attitude}}\\
	\midrule
	\texttt{experimental/11} & 1.02 & 0.84 & 1.65 & 0.91 & 99.71 & 91.20\\
	\texttt{experimental/12} & 1.17 & 1.08 & 0.86 & 0.87 & 99.72 & 100.00\\
	\bottomrule
 \end{tabular}
 \end{adjustbox}
\end{table}

The experimental evaluation demonstrates, on average, a higher attitude error than the synthetic evaluation case. In particular, for \texttt{experimental/11}, a spike in the initial \num{35} seconds of the sequence cause the error to surpass \SI{7.5}{\degree} which brings down the requirement compliance to \SI{91.2}{\percent}. This is due to the \gls{sv} travelling to a waypoint during the acquisition phase that is quite distinctive from the others present in the training data, making the berthing fixture appear in the top right corner of the \gls{fov} close to the image edge (Figure 27 a). However, the proposed training scheme which includes image augmentation prevents the error from diverging, and the estimate begins to recover after $\tau=\SI{35}{\second}$, reaching minimum values during and immediately before the final phase. In \texttt{experimental/12}, the attitude estimation error is bounded at \SI{2.5}{\degree}.
\section{Conclusions}

OOS is now becoming increasingly important and represents a significant cost saving measure, opening up a new global market. Whereas the latest OOS initiatives and demonstrators have focused on clear near-term commercial opportunities, such as life extension and end-of-life, longer-term OOS segments expected to emerge this decade such as refuelling are set to unravel novel and wider business opportunities, and have the potential to unlock new orbital ecosystems.

On this basis, City, University of London have developed OIBAR, a novel AI-based solution for space docking and refuelling applications consisting of the combination of two major components: a vision-based orbital relative navigation algorithm to safely approach and dock to the target vehicle; and an intelligent hardware mechanism achieving the mechanical docking and refuelling operation of the target. The present document reported the development and achievements of the OIBAR project, namely the design procedure of its key features adopted to tackle the problem, the modelling of the mechanism and software architecture, and the validation of the combined solution. Functional testing of the prototype was performed in laboratory using a 7-DOF robotic manipulator to simulate docking/berthing trajectories and a state-of-the-art Optitrack ground truth measurement system to assess the quality of the navigation solution. 

The proposed mechanism design sought to minimise the number of needed actuators in order to reduce the complexity and increase its reliability. The mechanism was first designed, analysed, and iterated upon in a 3D model simulation environment. Then, a prototype was built in a two-step process, first by resorting to 3D printing to assess basic mechanical functionality, and then through CNC machining for complete functional testing on the robotic testbed. Hardware validation demonstrated  that the berthing fixture was capable of capturing the end effector at a maximum relative velocity of 50 mm/s under combined yaw and pitch misalignments up to 10 deg, demonstrating resilience towards high angular misalignments. The registered contact force exerted on the end-effector did not reach 60 N for hard-docking; for soft-docking the reported value was reduced by more than half. Refuelling capability was implemented but not tested, and experiments will be carried out in the future.

A CNN-based direct VBS navigation algorithm was proposed to estimate the relative states between SV and TV to achieve docking. A MATLAB/Simulink simulator was developed to generate synthetic data intended to train and evaluate the solution. A benchmarking campaign was performed to assess the best architecture candidate. The final model reported average errors per trajectory of 1.19\% and 0.29 deg for range-normalised position and for attitude, respectively. The performance requirements were satisfied for nearly the whole length of the test sequences. The inclusion of BiLSTM-based recurrent layers was analysed but found not to improve the base CNN model.

Lastly, the combined solution was assessed through an integration testing campaign. The navigator was trained and tested on experimental data collected in laboratory using the mechanical docking prototype. The estimation results were in accordance with the synthetic dataset results, thus validating the findings. An overall increase in the mean attitude error, though, was registered, however, which was due to an increased variation in possible attitude states induced by the waypoint programming on the robotic manipulator. An enlargement of the training dataset poses is expected to further reduce the error.

\section*{Acknowledgements} 

\noindent This research has been supported by the UK Robotics and Artificial Intelligence (RAI) hub in Future AI and Robotics for Space (FAIR-SPACE: EP/R026092) and other sources.

\printbibliography

@inproceedings{benarroche2014atv,
	doi = {10.2514/6.2014-1665},
	year = 2014,
	month = {05},
	publisher = {American Institute of Aeronautics and Astronautics},
	author = {Patrice Benarroche and Martial Vanhove and Mauro Augelli},
	title = {{ATV Operations: from Demo Flight to Human Spaceflight Partner}},
	booktitle = {{SpaceOps} 2014 Conference}
}

@article{valmorbida2020calibration,
	doi = {10.1016/j.measurement.2019.107161},
	year = 2020,
	month = {2},
	publisher = {Elsevier {BV}},
	volume = {151},
	pages = {107161},
	author = {Andrea Valmorbida and Mattia Mazzucato and Marco Pertile},
	title = {Calibration procedures of a vision-based system for relative motion estimation between satellites flying in proximity},
	journal = {Measurement}
}

@inproceedings{chen2019satellite,
	doi = {10.1109/iccvw.2019.00343},
	year = 2019,
	month = {oct},
	publisher = {{IEEE}},
	author = {Bo Chen and Jiewei Cao and Alvaro Parra and Tat-Jun Chin},
	title = {{Satellite Pose Estimation with Deep Landmark Regression and Nonlinear Pose Refinement}},
	booktitle = {2019 {IEEE}/{CVF} International Conference on Computer Vision Workshop ({ICCVW})}
}

@inproceedings{cipolla2018multitask,
	doi = {10.1109/cvpr.2018.00781},
	year = 2018,
	month = {6},
	publisher = {{IEEE}},
	author = {Roberto Cipolla and Yarin Gal and Alex Kendall},
	title = {{Multi-task Learning Using Uncertainty to Weigh Losses for Scene Geometry and Semantics}},
	booktitle = {2018 {IEEE}/{CVF} Conference on Computer Vision and Pattern Recognition}
}

@inproceedings{deng2009imagenet,
	doi = {10.1109/cvpr.2009.5206848},
	year = 2009,
	month = {6},
	publisher = {{IEEE}},
	author = {Jia Deng and Wei Dong and Richard Socher and Li-Jia Li and  Kai Li and  Li Fei-Fei},
	title = {{ImageNet}: A large-scale hierarchical image database},
	booktitle = {2009 {IEEE} Conference on Computer Vision and Pattern Recognition}
}

@online{esa2021,
    author = {European~Space~Agency},
    title  = {{ESA invites ideas to open up in-orbit servicing market}},
    date   = {2021-04-01},
    url    = {https://www.esa.int/Safety_Security/Clean_Space/ESA_invites_ideas_to_open_up_in-orbit_servicing_market}
}

@inbook{fehse2003sensors,
    place={Cambridge},
    series={Cambridge Aerospace Series},
    title={Sensors for rendezvous navigation},
    doi={10.1017/CBO9780511543388.008},
    booktitle={{Automated Rendezvous and Docking of Spacecraft}},
    publisher={Cambridge University Press},
    author={Fehse, Wigbert}, year={2003},
    pages={218–282},
    collection={Cambridge Aerospace Series}
}

@InProceedings{graves2005bidirectional,
author="Graves, Alex
and Fern{\'a}ndez, Santiago
and Schmidhuber, J{\"u}rgen",
editor="Duch, W{\l}odzis{\l}aw
and Kacprzyk, Janusz
and Oja, Erkki
and Zadro{\.{z}}ny, S{\l}awomir",
title={{Bidirectional LSTM Networks for Improved Phoneme Classification and Recognition}},
booktitle="Artificial Neural Networks: Formal Models and Their Applications -- ICANN 2005",
year="2005",
publisher="Springer Berlin Heidelberg",
address="Berlin, Heidelberg",
pages="799--804",
isbn="978-3-540-28756-8",
doi = {10.1007/11550907_126}
}

@misc{he2015deep,
  doi = {10.48550/ARXIV.1512.03385},
  author = {He, Kaiming and Zhang, Xiangyu and Ren, Shaoqing and Sun, Jian},
  title = {Deep Residual Learning for Image Recognition},
  publisher = {arXiv},
  year = {2015},
  copyright = {arXiv.org perpetual, non-exclusive license}
}

@misc{hinton2012improving,
  doi = {10.48550/ARXIV.1207.0580},
  author = {Hinton, Geoffrey E. and Srivastava, Nitish and Krizhevsky, Alex and Sutskever, Ilya and Salakhutdinov, Ruslan R.},
  title = {{Improving neural networks by preventing co-adaptation of feature detectors}},
  publisher = {arXiv},
  year = {2012},
  copyright = {arXiv.org perpetual, non-exclusive license}
}

@misc{iandola2016squeezenet,
  doi = {10.48550/ARXIV.1602.07360},
  author = {Iandola, Forrest N. and Han, Song and Moskewicz, Matthew W. and Ashraf, Khalid and Dally, William J. and Keutzer, Kurt},
  title = {{SqueezeNet: AlexNet-level accuracy with 50x fewer parameters and <0.5 MB model size}},
  publisher = {arXiv},
  year = {2016},
  copyright = {arXiv.org perpetual, non-exclusive license}
}

@inproceedings{lei2023novel,
	year = 2023,
	month = {1},
	publisher = {American Institute of Aeronautics and Astronautics},
	author = {Lei He and Duarte Rondao and Nabil Aouf},
	title = {{A Novel Mechanism for Orbital AI-based Autonomous Refuelling}},
	booktitle = {2023 {AIAA} Guidance, Navigation, and Control Conference},
	location = {National Harbor, MD},
	note = {under submission}
}

@misc{kingma2014adam,
  doi = {10.48550/ARXIV.1412.6980},
  author = {Kingma, Diederik P. and Ba, Jimmy},
  title = {{Adam: A Method for Stochastic Optimization}},
  publisher = {arXiv},
  year = {2014},
  copyright = {arXiv.org perpetual, non-exclusive license}
}

@article{kisantal2020satellite,
	doi = {10.1109/taes.2020.2989063},
	year = 2020,
	month = {10},
	publisher = {Institute of Electrical and Electronics Engineers ({IEEE})},
	volume = {56},
	number = {5},
	pages = {4083--4098},
	author = {Mate Kisantal and Sumant Sharma and Tae Ha Park and Dario Izzo and Marcus Martens and Simone D{\textquotesingle}Amico},
	title = {Satellite Pose Estimation Challenge: Dataset, Competition Design, and Results},
	journal = {{IEEE} Transactions on Aerospace and Electronic Systems}
}

@article{krizhevsky2017imagenet,
	doi = {10.1145/3065386},
	year = 2017,
	month = {5},
	publisher = {Association for Computing Machinery ({ACM})},
	volume = {60},
	number = {6},
	pages = {84--90},
	author = {Alex Krizhevsky and Ilya Sutskever and Geoffrey E. Hinton},
	title = {{ImageNet} classification with deep convolutional neural networks},
	journal = {Communications of the {ACM}}
}

@book{markley2014fundamentals,
	doi = {10.1007/978-1-4939-0802-8},
	year = 2014,
	publisher = {Springer New York},
	author = {F. Landis Markley and John L. Crassidis},
	title = {{Fundamentals of Spacecraft Attitude Determination and Control}}
}

@conference{metcalfe2014robotic,
  title={{Robotic Refuelling Mission: Demonstrating Satellite Refuelling Technology on Board the ISS}},
  author={Metcalfe, Laurie and Hillebrandt, Tara },
  booktitle={Proceedings of 12\textsuperscript{th} International Symposium on Artificial Intelligence, Robotics and Automation in Space (i-SAIRAS)},
  address = {Montréal, Canada},
  year={2014},
  organization = {European Space Agency}
}

@online{northrop2020,
    author = {Cox, Vicki},
    title  = {{Northrop Grumman’s Wholly Owned Subsidiary, SpaceLogistics, Selected by DARPA as Commercial Partner for Robotic Servicing Mission}},
    date   = {2020-03-04},
    url    = {https://news.northropgrumman.com/news/releases/northrop-grummans-wholly-owned-subsidiary-spacelogistics-selected-by-darpa-as-commercial-partner-for-robotic-servicing-mission}
}

@inproceedings{proencca2020deep,
  title={Deep learning for spacecraft pose estimation from photorealistic rendering},
  author={Proen{\c{c}}a, Pedro F and Gao, Yang},
  booktitle={2020 IEEE International Conference on Robotics and Automation (ICRA)},
  pages={6007--6013},
  year={2020},
  organization={IEEE}
}

@inproceedings{redmon2017yolo9000,
	doi = {10.1109/cvpr.2017.690},
	year = 2017,
	month = {7},
	publisher = {{IEEE}},
	author = {Joseph Redmon and Ali Farhadi},
	title = {{YOLO9000: Better, Faster, Stronger}},
	booktitle = {2017 {IEEE} Conference on Computer Vision and Pattern Recognition ({CVPR})}
}

@misc{redmon2018yolov3,
      title={{YOLOv3: An Incremental Improvement}}, 
      author={Joseph Redmon and Ali Farhadi},
      year={2018},
      eprint={1804.02767},
      archivePrefix={arXiv},
      primaryClass={cs.CV}
}

@article{rondao2021chinet,
	doi = {10.1109/taes.2022.3193085},
	url = {https://doi.org/10.1109%2Ftaes.2022.3193085},
	year = 2022,
	publisher = {Institute of Electrical and Electronics Engineers ({IEEE})},
	pages = {1--13},
	author = {Duarte Rondao and Nabil Aouf and Mark A. Richardson},
	title = {{ChiNet: Deep Recurrent Convolutional Learning for Multimodal Spacecraft Pose Estimation}},
	journal = {{IEEE} Transactions on Aerospace and Electronic Systems}
}

@phdthesis{rondao2021multimodal,
  author       = {Duarte Rondao}, 
  title        = {{Multimodal Navigation for Accurate Rendezvous Missions}},
  school       = {Cranfield University},
  year         = 2021,
  note         = {Available from: \url{https://ethos.bl.uk/OrderDetails.do?uin=uk.bl.ethos.883372}}
}

@article{rondao2021robust,
	doi = {10.2514/1.g004794},
	year = 2021,
	month = {6},
	publisher = {American Institute of Aeronautics and Astronautics ({AIAA})},
	volume = {44},
	number = {6},
	pages = {1157--1182},
	author = {Duarte Rondao and Nabil Aouf and Mark A. Richardson and Vincent Dubanchet},
	title = {{Robust On-Manifold Optimization for Uncooperative Space Relative Navigation with a Single Camera}},
	journal = {Journal of Guidance, Control, and Dynamics}
}

@inproceedings{smith2017cyclical,
	doi = {10.1109/wacv.2017.58},
	year = 2017,
	month = {3},
	publisher = {{IEEE}},
	author = {Leslie N. Smith},
	title = {{Cyclical Learning Rates for Training Neural Networks}},
	booktitle = {2017 {IEEE} Winter Conference on Applications of Computer Vision ({WACV})}
}

@article{song2022deep,
	doi = {10.1016/j.actaastro.2021.10.025},
	year = 2022,
	month = {2},
	publisher = {Elsevier {BV}},
	volume = {191},
	pages = {22--40},
	author = {Jianing Song and Duarte Rondao and Nabil Aouf},
	title = {Deep learning-based spacecraft relative navigation methods: A survey},
	journal = {Acta Astronautica}
}

@inproceedings{szegedy2015going,
	doi = {10.1109/cvpr.2015.7298594},
	year = 2015,
	month = {jun},
	publisher = {{IEEE}},
	author = {Christian Szegedy and  Wei Liu and  Yangqing Jia and Pierre Sermanet and Scott Reed and Dragomir Anguelov and Dumitru Erhan and Vincent Vanhoucke and Andrew Rabinovich},
	title = {Going deeper with convolutions},
	booktitle = {2015 {IEEE} Conference on Computer Vision and Pattern Recognition ({CVPR})}
}

@Inbook{szeliski2022computer,
    author="Szeliski, Richard",
    title={{Structure from motion and SLAM}},
    bookTitle="Computer Vision: Algorithms and Applications",
    year="2022",
    publisher="Springer International Publishing",
    address="Cham",
    pages="543--594",
doi="10.1007/978-3-030-34372-9_11",
}

@Inbook{wie2014attitude,
    author="Wie, Bong
    and Lappas, Vaios
    and Gil-Fern{\'a}ndez, Jes{\'u}s",
    editor="Macdonald, Malcolm
    and Badescu, Viorel",
    title="Attitude and Orbit Control Systems",
    bookTitle="The International Handbook of Space Technology",
    year="2014",
    publisher="Springer Berlin Heidelberg",
    address="Berlin, Heidelberg",
    pages="323--369",
    isbn="978-3-642-41101-4",
    doi="10.1007/978-3-642-41101-4_12",
}

@techreport{uksa2021,
  author      = {{Astroscale, Fair-Space, and Catapult~Space~Applications}},
  title       = {UK In-Orbit Servicing Capability},
  subtitle    = {A Platform for Growth},
  institution = {UK Space Agency},
  year        = {2021-05},
  type        = {Technical Report},
  address     = {Swindon, UK},
  url         = {https://sa.catapult.org.uk/wp-content/uploads/2021/05/Catapult-Astroscale-Fairspace-Platform-for-Growth-report-final-27-05-21.pdf}
  }

@misc{zhou2018continuity,
  doi = {10.48550/ARXIV.1812.07035},
  author = {Zhou, Yi and Barnes, Connelly and Lu, Jingwan and Yang, Jimei and Li, Hao},
  title = {On the Continuity of Rotation Representations in Neural Networks},
  publisher = {arXiv},
  year = {2018},
  copyright = {arXiv.org perpetual, non-exclusive license}
}

\end{document}